\documentclass[twoside]{article}
\usepackage{ecj,palatino,epsfig,latexsym,natbib, amsmath}
\usepackage{bbm}
\usepackage{txfonts}
\usepackage{float}
\usepackage{bm}

\usepackage{titlesec}
\newtheorem{definition}{Definition}
\DeclareMathOperator\supp{supp}

\begin{document}

\ecjHeader{x}{x}{xxx-xxx}{201X}{Quality Diversity for Grasping in Robotics}{J Huber, F Hélénon, M Coninx, F Ben Amar, S Doncieux}
\title{\bf Quality Diversity under Sparse Interaction and Sparse Reward: Application to Grasping in Robotics}  

\author{\name{\bf Johann Huber} \hfill \addr{johann.huber@isir.upmc.fr}\\ 
        \addr{Sorbonne Université, CNRS, ISIR\thanks{ \hspace{1mm} Institut des Systèmes Intelligents et de Robotique} , Paris, 75005, France}
\AND
       \name{\bf François Helenon} \hfill \addr{francois.helenon@isir.upmc.fr}\\
        \addr{Sorbonne Université, CNRS, ISIR\footnotemark[1] , Paris, 75005, France}
\AND
       \name{\bf Miranda Coninx} \hfill \addr{miranda.coninx@isir.upmc.fr}\\
        \addr{Sorbonne Université, CNRS, ISIR\footnotemark[1] , Paris, 75005, France}
\AND
       \name{\bf Faïz Ben Amar} \hfill \addr{faiz.ben-amar@isir.upmc.fr}\\
        \addr{Sorbonne Université, CNRS, ISIR\footnotemark[1] , Paris, 75005, France}
\AND
       \name{\bf Stéphane Doncieux} \hfill \addr{stephane.doncieux@isir.upmc.fr}\\
        \addr{Sorbonne Université, CNRS, ISIR\footnotemark[1] , Paris, 75005, France}
}

\maketitle

% TeX files inputs

\begin{abstract}

% ~200 mots: ok

Quality-Diversity (QD) methods are algorithms that aim to generate a set of diverse and high-performing solutions to a given problem. Originally developed for evolutionary robotics, most QD studies are conducted on a limited set of domains – mainly applied to locomotion, where the fitness and the behavior signal are dense. Grasping is a crucial task for manipulation in robotics. Despite the efforts of many research communities, this task is yet to be solved. Grasping cumulates unprecedented challenges in QD literature: it suffers from reward sparsity, behavioral sparsity, and behavior space misalignment. The present work studies how QD can address grasping. Experiments have been conducted on 15 different methods on 10 grasping domains, corresponding to 2 different robot-gripper setups and 5 standard objects. An evaluation framework that distinguishes the evaluation of an algorithm from its internal components has also been proposed for a fair comparison. The obtained results show that MAP-Elites variants that select successful solutions in priority outperform all the compared methods on the studied metrics by a large margin. We also found experimental evidence that sparse interaction can lead to deceptive novelty. To our knowledge, the ability to efficiently produce examples of grasping trajectories demonstrated in this work has no precedent in the literature.

\end{abstract}

% DISTINGUER ALIGNEMENT COMPORTEMENTAL 
% et nouvelle notion: completude ou densité comportementale


%%%%%%%%%%%%%%%%%%%%%%%%%%%%%%%%%%%%%%%%%%%%%%%%%%%%%%%%%%%%%%%%%%%%%%%
%                        1) Introduction
%%%%%%%%%%%%%%%%%%%%%%%%%%%%%%%%%%%%%%%%%%%%%%%%%%%%%%%%%%%%%%%%%%%%%%%

\section{Introduction}
\label{sec:1_introduction}

Quality-Diversity (QD) methods are evolutionary algorithms that optimize both diversity and quality to generate large repertoires of high-performing solutions to a given problem (\cite{pugh2016quality}, \cite{cully2017quality}). This field produced significant results in evolutionary robotics, including recovery from injury (\cite{cully2015robots}), generation of adversarial objects for robotic grasping (\cite{morrison2020egad}), or morphological evolution (\cite{zardini2021seeking}). The recent rise of interest led to novel interactions between fields, with notable combinations of QD and Reinforcement-Learning (\cite{sigaud2022combining}) or with supervised learning (\cite{mace2023quality}). Numerous ideas are explored to further carry the field regarding the method (\cite{fontaine2021differentiable}, \cite{faldor2023map}) or trying to address more complex tasks (\cite{anne2023multi}, \cite{flageat2023uncertain}).

Interestingly, \textbf{most QD domains for evolutionary robotics are tasks in which the fitness and the behavioral functions deliver non-constant signals, making them exploitable}. Those domains usually involve navigation, where it is easy to design (\cite{faldor2023map}) or to automatically learn (\cite{paolo2020unsupervised}) a behavior space that expresses the agent's displacement. Even if the fitness function is usually orthogonal to the targeted task – e.g. energy minimization while trying to generate locomotion policies – the function is always defined such that the algorithm can continuously optimize both the diversity and the quality of the processed solutions.

Recently, \cite{paolo2021sparse} studied tasks submitted to sparse fitness, proposing a QD algorithm that optimizes quality despite the limited reward signal. However, all the considered tasks involve the navigation of an entity within the environment (the agent itself or a ball to push somewhere): a behavioral space defined as the key entity's last position gives the targeted task's complete information. \textbf{Such a behavioral characterization can be said \textit{complete}, as it provides the guarantee that the optimal solution will eventually be found} (NS assumption of uniform exploration (\cite{doncieux2019novelty})) \textbf{and that the algorithm can always rely on an exploitable behavioral signal throughout the evolutionary process} (see section \ref{sec:3_3_1_define_b_for_qd}). This work argues that \textbf{some challenging tasks like robotic manipulation ones cannot be addressed under such a reliable behavioral characterization}.

Grasping refers to making an agent pick an object by applying forces and torques on its surface. Considered a prerequisite for many manipulation tasks (\cite{hodson2018gripping}), the sparsity of grasping's reward makes data-oriented approaches struggle in these domains. Despite efforts from many research communities, grasping is still only partially solved (\cite{zhang2022robotic}). An essential matter for solving grasping with learning methods is the ability to generate demonstrations that can bootstrap learning (\cite{wangdemograsp}, \cite{de2020learning}). \textbf{The present work shows that QD methods can be reliably leveraged to generate a large set of diverse high-performing solutions that fulfill this need for high-quality data}. 

While defining the proper behavioral characterization is not trivial, we here consider the first Cartesian position of the end effector when touching the object for the first time. By randomly initializing robotic arm policies, a significant part of the trajectories do not even touch the object. Therefore, those evaluations do not provide any exploitable behavioral information. Grasping is therefore not only submitted to sparse reward but also to \textit{sparse interactions}. Plus, a behavioral characterization that guarantees that successful solutions will eventually be found is hardly designable. To our knowledge, \textbf{this work is the first that demonstrates QD algorithms capabilities to solve sparse fitness and sparse interaction problems without relying on an aligned behavior space.}

This paper aims to \textbf{study how the QD literature can scale up to sparse reward and sparse interaction tasks}, by comprehensively studying \textbf{how QD can be applied to the yet unsolved task of grasping}. This investigation raises many questions on QD, including the role of the behavioral characterization, the taxonomy of QD methods, and the evaluation procedure. It also leads to key insights on how QD methods perform on tasks submitted to sparse rewards and interaction, and bring to light the critical algorithmic components that make a QD method work in this context. Finally, this study led to insights into how QD methods can do efficient exploration in sparse interaction problems, showing that contrary to tasks where the behavior function is dense, novelty-driven approaches have poor exploration capabilities on sparse interaction problems.

Our contributions are the following:

\begin{itemize}
    \item We propose a taxonomy to avoid ambiguities when talking about QD methods, especially regarding NS-related methods;
    \item We discuss the role of the behavioral characterization through the notions of \textit{behavioral alignment}, \textit{density of the behavioral function}, \textit{behavioral completeness} and \textit{driving/describing behavior space};
    \item We introduce a simple framework that distinguishes the evaluation of a QD method from its internal components;
    \item We show that a simple variant of MAP-Elites consistently dominates state-of-the-art QD methods on the considered metrics, demonstrating capabilities to generate a large set of diverse and high-performing solutions on grasping – despite this task challenges: sparse fitness, sparse interaction, and behavior misalignment;
    \item We investigate the impact of the behavioral sparsity on QD methods performances, obtaining empirical evidence that sparse interaction can lead to deceptive novelty.
\end{itemize}

The code is available on Github\footnote{https://github.com/Johann-Huber/qd\_grasp}. We believe these results will open the way to apply QD on more complex tasks related to robotic manipulation, eventually solving problems that cannot easily be tackled with learning methods from other fields. The experimental results demonstrated here show that \textbf{QD methods can efficiently be leveraged to generate grasping trajectories of different fitnesses}. Such data could be used to bootstrap learning strategies of any kind. Generating grasping demonstrations is a \textbf{key matter to solve this task} (\cite{wangdemograsp}, \cite{de2020learning}). To our knowledge, \textbf{no method in the literature is able to easily produce examples of grasping trajectories on different robots and objects as shown in this work}.

%%%%%%%%%%%%%%%%%%%%%%%%%%%%%%%%%%%%%%%%%%%%%%%%%%%%%%%%%%%%%%%%%%%%%%%
%                        2) Related works
%%%%%%%%%%%%%%%%%%%%%%%%%%%%%%%%%%%%%%%%%%%%%%%%%%%%%%%%%%%%%%%%%%%%%%%

\section{Related works}
\label{sec:2_related_works}

%=====================================================================%
%                   2.1) Quality Diversity
%=====================================================================%

\subsection{Quality diversity}
\label{sec:2_1_qd}

While standard optimization approaches search for the extremum solution to a single-objective solution, Quality-Diversity (QD) algorithms aim to generate a set of diverse and high-performing solutions. Those methods lead to application in many fields, including image generation (\cite{fontaine2023covariance}), discovery of drugs (\cite{verhellen2020illuminating}), or engineering optimization (\cite{gaier2018data}). QD methods rely on a behavioral characterization to compare the evaluated solutions for a given task, allowing to maintain diversity along with the optimization of a quality criterion.

QD methods emerged through two seminal works, NSLC and MAP-Elites. NSLC (\cite{lehman2011evolving}) is a population-based evolutionary method that adds pressure toward the most novel and high-performing individuals through a Pareto-front selection. It has first been introduced as an extension of Novelty Search (NS) (\cite{lehman2011abandoning}) – an approach that replaces the quality-guided optimization process with a novelty-guided one. MAP-Elites (ME) (\cite{mouret2015illuminating}) is the second seminal QD method. It relies on a structured container that keeps the best previously generated solutions for different behavioral niches. Almost all QD algorithms derivate from those two pioneer methods. However, MAP-Elites-based algorithms seem to be the most popular ones: most of the current state-of-the-art methods for rapid illumination of a behavior space are more or less complex variants of ME (\cite{fontaine2023covariance}, \cite{mace2023quality}, \cite{faldor2023map}).

Most of the QD works in robotics are actually applied to locomotion (\cite{lehman2011evolving}, \cite{mace2023quality}, \cite{faldor2023map}, \cite{zardini2021seeking}). In these domains, it is easy to design a behavioral characterization that expresses the agent's displacement, such that the exploration of this behavior space will eventually lead to the optimal solution (\cite{lehman2011abandoning}, \cite{paolo2021sparse}). The present work aims to show that QD can efficiently be applied to more complex tasks like grasping, despite the involved challenges: sparse fitness, sparse interaction, and misaligned behavior space (see section \ref{sec:3_3_1_define_b_for_qd}).

The present work calls \textit{complete} a behavioral characterization that allows an easy exploration by guaranteeing that a successful solution will eventually be found through the use of a non-constant
behavioral signal (see section \ref{sec:3_3_1_4_behavioral_completeness}). This paper shows that \textbf{QD methods can efficiently be applied on more complex tasks like robot manipulation, in which the algorithm cannot rely on a complete behavioral characterization.}

%=====================================================================%
%                   2.2) QD for hard exploration problems
%=====================================================================%

\subsection{QD for hard exploration problems}
\label{sec:2_2_qd_hard_explore}

%--------------------------------------------------------------------%
%               2.2.1) NS and ME divergence
%--------------------------------------------------------------------%

\begin{figure}[t]
\begin{center}
\centerline{
 \includegraphics[width=0.9\textwidth]{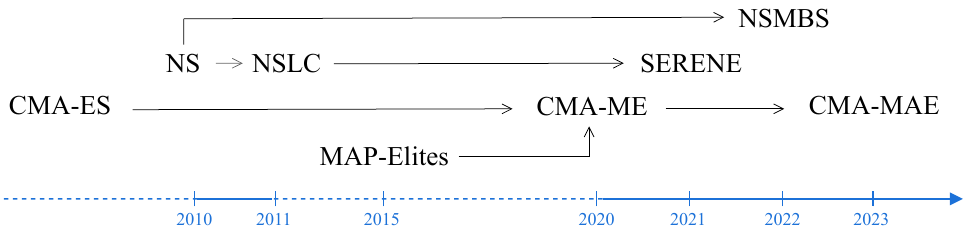}
}
\end{center}
\caption{\textbf{Timeline of compared methods.} The literature is split into NS-derivated methods and MAP-Elites-derivated methods. In this work, we compare the 3 most promising methods for rapid illumination of a behavior space in sparse reward and behavior (SERENE, NSMBS, and CMA-MAE) with 12 other methods mainly derived from NS and ME. To fairly compare them, we proposed an evaluation framework in which the collected metrics are not dependent on the algorithms' components.}
\label{fig:timeline_compared_methods}
\end{figure}

\subsubsection{NS and ME divergence}

Novelty Search has been introduced as a promising approach to address sparse or deceptive reward problems (\cite{lehman2011abandoning}. This seminal paper led to the emergence of Quality Diversity, as NSLC, the first QD method, derivates from NS. Interestingly, the QD field has led to two research branches: \textbf{NS-based} and \textbf{ME-based works} (Figure \ref{fig:timeline_compared_methods}). NS and ME-based methods are rarely compared in works involving one or the other family (\cite{fontaine2023covariance}, \cite{kim2021exploration}). All those methods involve similar properties and mechanisms; we argue that \textbf{comparing them might lead QD practitioners to insightful results}.

%--------------------------------------------------------------------%
%               2.2.2) QD for sparse reward domains
%--------------------------------------------------------------------%

\subsubsection{QD for sparse reward domains}

Recent works have studied how QD methods could address domains submitted to sparse rewards: SERENE (\cite{paolo2021sparse}) proposes a new approach to optimize a real-valued fitness function in the sparse context; NSMBS (\cite{morel2022automatic}) explores multiple behavior spaces to generate grasping trajectories; CMA-MAE (\cite{fontaine2023covariance}) claims to fix the limitations of CMA-ME (\cite{fontaine2020covariance}) on flat fitness landscapes, and reported state-of-the-art results on challenging tasks. Those methods can be considered as the most promising algorithms for addressing the task of grasping, which is submitted to sparse reward: SERENE is the only QD method that explicitly does Rapid Illumination of a Behavior Space (RIBS) in sparse reward, NSMBS is the only QD method that demonstrated results on grasping, and CMA-MAE is a state-of-the-art method for doing RIBS, getting specific algorithmic mechanism to be more robust to sparse reward domains. \textbf{The present works compared these 3 methods with 12 other algorithms to identify the best-performing approach for doing RIBS on grasping.}

%=====================================================================%
%                   2.3) Grasping
%=====================================================================%

\subsection{Grasping} 
\label{sec:2_3_graspping}
Grasping refers to making an agent solidarize its end effector with an object by applying forces and torques on its surface. This task is of great interest to robotics and artificial intelligence research communities, as it is considered a prerequisite for many manipulation tasks (\cite{hodson2018gripping}). After the early ages of analytical-based methods (\cite{nguyen1988constructing}), data-driven approaches have dominated the literature on grasping in robotics since the beginning of the 21st century (\cite{zhang2022robotic}). Despite the involved research efforts, \textbf{grasping is still partially solved}: the reward sparsity of grasping makes it very challenging for learning methods to generate data to bootstrap learning from. To increase the chances of success of random movements, most of the approaches constrain the operational space to top-down movements (\cite{yang2023pave}) or are limited to parallel grippers (\cite{fang2020graspnet}). As the self-supervised acquisition of data is very expensive (\cite{levine2018learning}), most of the recent works on grasping rely on human-provided demonstrations (\cite{wangdemograsp}, \cite{mosbach2023learning}). These promising results are at the cost of human-provided demonstrations, which is time expensive. The resulting grasping policies' adaptation capabilities are thus limited by the provided examples. 

\textbf{The acquisition of demonstrations that can bootstrap learning is thus a key matter for solving grasping.} Ideally, those demonstrations should be acquired in simulation only (to avoid the issues raised by long-term self-supervised learning on real robots (\cite{levine2018learning})), generated with limited human intervention, suited to many grasping scenes (robot, end effector, and objects), and diverse enough to foster generalization capabilities of the learned policies. In this work, \textbf{we leverage QD algorithms to generate large sets of diverse grasping trajectories}. The presented results show that a QD method can successfully generate grasping datasets for \textbf{different end effectors and objects}. Plus, the optimized fitness functions \textbf{associate to each generated grasp a quality label} that can be straightforwardly used for training.

%%%%%%%%%%%%%%%%%%%%%%%%%%%%%%%%%%%%%%%%%%%%%%%%%%%%%%%%%%%%%%%%%%%%%%%
%                           3) Problem
%%%%%%%%%%%%%%%%%%%%%%%%%%%%%%%%%%%%%%%%%%%%%%%%%%%%%%%%%%%%%%%%%%%%%%%

\section{Problem}
\label{sec:3_problem}

%=====================================================================%
%                   3.1) Notations
%=====================================================================%

\begin{figure}[t]
\begin{center}
\centerline{
 \includegraphics[width=0.7\textwidth]{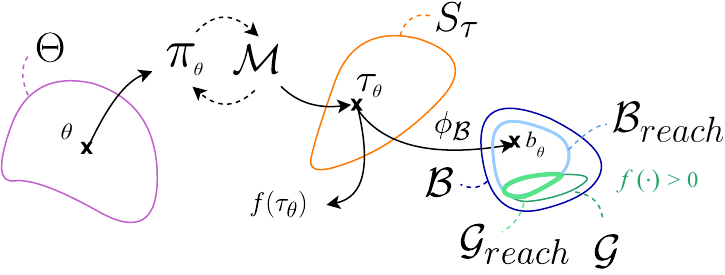}
}
\end{center}
\caption{\textbf{Notations overview.} Each \textit{individual} $\theta\in\Theta$ define a \textit{policy} $\pi_\theta$. This policy allows an agent to interact with an \textit{environment} corresponding to a Markov Decision Process $\mathcal{M}$. The sequence of states and actions obtained after an \textit{evaluation} on $T$ steps is called the resulting \textit{trajectory} $\tau_\theta$. This high dimensional vector is then projected in a behavior space $\mathcal{B}$ through a \textit{behavior function} $\phi_\mathcal{B}$, obtaining a \textit{behavior descriptor} $b_\theta$ in low dimensions. $\tau_\theta$ also allow the computation of the fitness $f(\tau_\theta)$. The QD algorithm conducts its selection-mutation process using the resulting $f(\tau_\theta)$ and $b_\theta$. The goal space $\mathcal{G}$ is the subset of $\mathcal{B}$ in which the corresponding trajectories have led to a non-null fitness.}
\label{fig:overall_notations}
\end{figure}

\subsection{Notations}
\label{sec:3_1_notations}
%--------------------------

This section introduces the notations used throughout this paper. The following subsections describe the QD notations background (section \ref{sec:3_1_1_qd}), the notion of fitness sparsity (section \ref{sec:3_1_3_fitness_sparsity}), behavioral sparsity (section \ref{sec:3_1_4_behavior_sparsity_and_elig}), and the notations related to policy learning applied to robotics (section \ref{sec:3_1_5_policy_learning_robotics}). Figure \ref{fig:overall_notations} overviews the notations.

%--------------------------------------------------------------------%
%                           3.1.1) Background
%--------------------------------------------------------------------%

\subsubsection{Background}
\label{sec:3_1_1_qd}

This work relies on QD standard notations (\cite{cully2022quality}). Let $\Theta \subseteq \mathbb{R}^{n_\theta}$ be the \textit{parameters space}, $\theta \in \Theta$ an \textit{individual} (also referred as a \textit{genome} or a \textit{solution}). Let $\mathcal{B} \subseteq \mathbb{R}^{n_b}$ be the \textit{behavior space}, We note $\phi_{\mathcal{B}}:\Theta \rightarrow \mathcal{B}$ the \textit{behavior function}, such that $b_\theta = \phi_{\mathcal{B}}(\theta)$ is the \textit{behavior descriptor} of $\theta$. Let $f:\Theta\rightarrow \mathbb{R}$ be the \textit{fitness function}, $d_{\mathcal{B}}:\mathcal{B}^2 \rightarrow \mathbb{R}$ a distance function within $\mathcal{B}$. We aim to generate an \textit{archive} $A$ defined as:
\begin{equation}
\left\{\begin{matrix}
\forall b \in \mathcal{B}_{reach}, \, \exists \theta \in A, \, d_{\mathcal{B}}(\phi_{\mathcal{B}}(\theta), b) < \epsilon \\
\forall \theta' \in A, \, \theta' =  \text{argmax}_{\theta\in N(b_{\theta'})}f(\theta) 
\end{matrix}\right.
\label{eq:qd_background}
\end{equation}
where $\mathcal{B}_{reach} \subseteq \mathcal{B}$ is the \textit{space of reachable behaviors}, $\epsilon\in\mathbb{R}^{+*}$ is a small value that defines the density of $\mathcal{B}_{reach}$ paving, and $N(b_{\theta'})= \{ \theta \mid neighbor_{d_{\mathcal{B}}}(b_\theta, b_{\theta'}) \}$ is the set of solutions for which the projection in $\mathcal{B}$ are close to each others. $\phi_{\mathcal{B}}$ is supposed deterministic. The function $neighbor_{d_{\mathcal{B}}}$ usually corresponds to a k-nearest neighbors algorithm (\cite{lehman2011abandoning}, \cite{lehman2011evolving}) that relies on $d_{\mathcal{B}}$. 

As the QD objective is usually to illuminate a behavioral space (i.e. fill $A$ with high-performing solutions), we can see those problems as looking for a way to explore $\mathcal{B}$ while sampling parameters from $\Theta$, considering that $\phi_{\mathcal{B}}$ is unknown and usually highly non-linear (\cite{doncieux2019novelty}). The exploration of $\mathcal{B}$ – and the concurrent optimization of $f$ – is carried through the evaluations of a \textit{domain} (or \textit{environment}). This step is described in subsection \ref{sec:3_1_5_policy_learning_robotics} through the prism of policy learning in robotics.

A large variety of QD methods exists in the literature (section \ref{sec:2_1_qd}), for which the archive $A$ can be of different nature. $A$ can be structured (\cite{mouret2015illuminating}), unstructured (\cite{lehman2011evolving}), composed of several sub-spaces (\cite{morel2022automatic}), or with depth (\cite{flageat2020fast}). There can also be variants without any archive (\cite{salehi2021br}). As we want to consider several kinds of QD algorithms, we here distinguish the running archive $A$ from the \textit{outcome archive} $A_o$. $A$ is used by the algorithm during the evolutionary proccess, while $A_o$ is an external archive used to analyse the outcome of the algorithm. Section \ref{sec:4_3_1_algo_output} describes how $A$ and $A_o$ are distinguished in practice.

%--------------------------------------------------------------------%
%                       3.1.2) Fitness sparsity
%--------------------------------------------------------------------%

\subsubsection{Fitness sparsity}
\label{sec:3_1_3_fitness_sparsity}

This work focuses on the sparse fitness context. In grasping, most of the evaluated $\theta_i$ result in $f(\theta_i)=0$. Let $f_c:\Theta\rightarrow \left\{0, 1\right\}$ be the sparse \textit{success criterion}, such that $f_c(\theta)=\mathbbm{1}_{f(\theta)>0}$. In this context, the actual output of QD is a \textit{success archive} $A_s$ defined as $A_s = \left\{ \theta \in A_o \mid f_c(\theta)=1 \right\}$. Thus, the space to explore is the \textit{goal space} $\mathcal{G}$ defined as $\mathcal{G} = \left\{ b_\theta \in \mathcal{B} \mid f(\theta)>0 \right\}$, and more accurately the \textit{space of reachable goals} $\mathcal{G}_{reach} = \left\{ b_\theta \in \mathcal{B}_{reach} \mid f(\theta)>0 \right\}$.

%--------------------------------------------------------------------%
%               3.1.3) Behavior sparsity and eligibility
%--------------------------------------------------------------------%

\subsubsection{Behavior sparsity and eligibility}
\label{sec:3_1_4_behavior_sparsity_and_elig}

This paper argues that applying QD to grasping requires facing a sparse interaction problem: $\exists \theta \in \Theta$ such that $\phi_{\mathcal{B}}(\theta)$ is not defined. We say here that $\theta$ results in a \textit{non-eligible} behavior descriptor\footnote{In practice, many QD algorithms require each individual to have a defined behavior descriptor. To avoid this issue, we set non-eligible descriptors to a vector of 0. For that reason, we can refer to the search space region in which the resulting behavior is eligible as the \textit{support} of $\phi_{\mathcal{B}}$.}. The set of \textit{eligible} behavior descriptors $\mathcal{B}_{elig}$ are defined as $\mathcal{B}_{elig} = \left\{ b \mid \exists \theta \in \Theta, b = \phi_{\mathcal{B}}(\theta) \right\}$. In this study, $\mathcal{B}_{elig} = \mathcal{B}_{reach}$ (see section \ref{sec:3_3_2_define_b_for_grasping}). 

%What differs from previous $\mathcal{B}_{reach}$ usage (\cite{doncieux2019novelty}) is that $\forall \theta \in \Theta, \exists b \in \mathcal{B}_{reach}, b=\phi_{\mathcal{B}}(\theta)$ while $\exists \theta \in \Theta, \phi_{\mathcal{B}}(\theta) = \o $.  
% À préciser au besoin

%--------------------------------------------------------------------%
%              3.1.4) Policy learning in Robotics
%--------------------------------------------------------------------%

\subsubsection{Policy learning in Robotics}
\label{sec:3_1_5_policy_learning_robotics}

We evaluate a policy $\pi_{\theta}$ on the Markov Decision Process $\mathcal{M}$, corresponding to the task's environment. We then obtain a trajectory $\tau_\theta \in S_\tau$, defined as a sequence of states and actions for each evaluation step along the \textit{episode}. $S_{\tau}$ is usually in high dimension. We then project $\tau_\theta$ from $S_{\tau}$ to a space $\mathcal{B}$ s.t. $dim({\mathcal{B}}) << dim(S_{\tau})$, and uses $N$ to compare resulting behaviors. We here consider a fixed initial state $s_0$ and a deterministic $\mathcal{M}$. 

In practice, the behavior function $\phi_{\mathcal{B}}$ is the result of the interaction of $\pi_\theta$ with $\mathcal{M}$, and the projection in $\mathcal{B}$ of the resulting trajectory $\tau_\theta$. To better match the conditions of present work's experiments, we will consider that the evaluation of a policy $\pi_\theta$ lead to a point $\tau_\theta$ in the trajectory space $S_\tau$, and the above-defined fitness functions project elements of $S_\tau$ to their respective space: $\phi_{\mathcal{B}}:S_\tau \rightarrow \mathcal{B}$  and $f:S_\tau \rightarrow \mathbb{R}$.

%=====================================================================%
%                   3.2) Taxonomy of QD methods
%=====================================================================%

\subsection{Taxonomy of QD methods}
\label{sec:3_2_taxonomy_of_qd_methods}

Quality Diversity (QD) and Novelty Search (NS) are concepts that describe specific methods and imply the emergence of specific addressable problems. From an algorithmic point of view, there is a clear difference between those two families of methods: NS does not rely on a fitness function – while QD does.

Nevertheless, those notions might easily be mixed up. Many works include vanilla NS among tested methods that derivate from NSLC and MAP-Elites (\cite{pugh2016quality}, \cite{paolo2021sparse}), while some others introduce as NS-related some methods that match the abovementioned definition of QD (\cite{kim2021exploration}, \cite{morel2022automatic}). The commonly shared definition of QD – an algorithm that maintains diversity and optimizes a local quality criterion – is itself not restrictive enough to prevent NS from being considered as a specific instance of QD algorithm\footnote{NS novelty can be actually be considered as a dynamic quality criterion the algorithm is optimizing by maintaining a diversity of solutions throughout the evolutionary process.}.

\textbf{We believe that some key insights can result from an exhaustive comparison of QD methods of different nature.} Such an approach requires to avoid the mentioned ambiguities. To make the analysis easier, \textbf{we propose a taxonomy that allows us to focus on each of the similarities and differences between those methods}. Actually, the ambiguities come from the usage of \textit{NS-based} and \textit{QD-based} methods for different purposes. Here is a list of the main reasons for the ambiguous usage of those notions and the alternative we propose to clarify them:

\begin{itemize}
    \item \textbf{Root algorithm.} The most well-spread misuse of language on that matter is to distinguish algorithms derivated from MAP-Elites (\cite{mouret2015illuminating}) and those derivated from Novelty Search (\cite{lehman2011abandoning}) by respectively calling them NS-based methods and QD-based methods. The implicit idea behind this shortcut is that those derivated algorithms share common properties of the root algorithm. The problem is that we cannot state if the shared property is the nature of the archive, the way the population is generated, or even the overall goal. In practice, many algorithms share common properties with NS (\cite{lehman2011evolving}, \cite{kim2021exploration}, \cite{morel2022automatic}) or with MAP-Elites (\cite{bruneton2019exploration}, \cite{nilsson2021policy}, \cite{mace2023quality}). We will thus refer to those families of methods as \textit{NS-derivated} and \textit{ME-derivated}, using any of the below-proposed distinctions as needed to avoid confusion.
    
    \item \textbf{Nature of the archive.} One might talk about NS-based methods for unstructured-archive-based ones, and QD-based methods for structured-archive based ones (\cite{cully2022quality}). The structured archive is here expressing the container based on a grid (\cite{mouret2015illuminating}) from those in which the novelty is computed with a nearest-neighbors approach (\cite{lehman2011abandoning}). Note that the introduction of CVT-MAP-Elites (\cite{vassiliades2017using}) and the notion of minimal novelty to add an individual into an archive (\cite{lehman2010revising}) makes those containers work similarly. To clarify this matter, we will always use the notions of \textit{UA-based} (Unstructured-Archive) and \textit{SA-based} (structured-archive) methods to distinguish them.
    
    \item \textbf{Population.} Considering that UA-based methods result from NSLC (\cite{lehman2011evolving}) and SA-based from MAP-Elites (\cite{mouret2015illuminating}), one might use the notion of NS-based methods for population-based one – in which a living population is maintained concurrently with a container – and QD-based methods for non-population-based one – in which offspring are generated from individuals directly sampled from the container. But nothing prevents NS-derivated methods from relying on the container to sample parents. An explicit distinction should thus be made between \textit{PP-based} methods (population-based) and \textit{NPP-based} methods (non-population-based).
    
    \item \textbf{Run goal.} Another point of ambiguity is the purpose of the algorithm execution itself. While the ultimate goal of QD methods is clearly established (i.e. generate a set of diverse and high-performing solutions – here referred to as \textit{Rapid Illumination of a Behavior Space (RIBS)}), NS-derivated methods aim to uniformly cover a behavior space (referred here as \textit{Behavior Space Coverage, BSC}) (\cite{wiegand2020objective}, \cite{doncieux2019novelty}) and by doing so, find an optimal solution to a given problem (\textit{single objective optimization, SOO}) (\cite{lehman2011abandoning}, \cite{shorten2014evolvable}).
\end{itemize} 

Finally, it is worth noting that the notion of Quality-Diversity becomes the most well-spread terminology when referring to the overall field. We will thus use \textit{QD} and \textit{QD-based} methods to talk about the field in general and distinguish it from other research perspectives on policy learning.

This work considers state-of-the-art QD methods to generate a diverse set of high-performing grasping trajectories. To do so, \textbf{we compare methods with any of the abovementioned mechanisms}, considering that \textbf{some could have been designed to address another primary purpose} but \textbf{might lead to promising results or properties on the considered task.}

%=====================================================================%
%           3.3) Behavioral characterization
%=====================================================================%

\subsection{Behavioral characterization}
\label{sec:3_2_bc_for_grasping}

Grasping is a challenging task (see section \ref{sec:2_3_graspping}) in which the definition of the behavior space $\mathcal{B}$ is not trivial. This section motivates the behavioral characterization used in this work experiment. At first are introduced the concepts and notions that can help to define a behavior space $\mathcal{B}$ for a new problem to address (section \ref{sec:3_3_1_define_b_for_qd}). These concepts are then applied to grasping, in which some facilitating behavioral hypotheses cannot be verified (section \ref{sec:3_3_2_define_b_for_grasping}).

%--------------------------------------------------------------------%
%               3.3.1) Defining B for a QD problem
%--------------------------------------------------------------------%

\subsubsection{Defining $\mathcal{B}$ for a QD problem}
\label{sec:3_3_1_define_b_for_qd}

This subsection provides definitions of key behavioral concepts. These notions allow to define a suited $\mathcal{B}$ for a given problem but also to stress the challenges caused by its behavioral characterization. At first are defined the notions of \textit{driving} and \textit{describing} $\mathcal{B}$ (section \ref{sec:3_3_1_1_driving_b_vs_describing_b}), followed by the notion of \textit{task alignment} (section \ref{sec:3_3_1_2_task_alignment}), then the \textit{density} of a behavioral function (section \ref{sec:3_3_1_3_behavioral_density}) and finally the notion of \textit{behavioral completeness} (section \ref{sec:3_3_1_4_behavioral_completeness}).

%We first discuss the roles of $\mathcal{B}$ in QD methods (section \ref{sec:3_2_1_describing_b_vs_driving_b}). We then define the notion \textit{alignment} between $\mathcal{B}$ and a task – showing that contrary to most of the standard QD domains, we cannot easily design $\mathcal{B}$ to make it aligned with grasping (section \ref{sec:3_2_2_task_alignment}). We finally describe the chosen $\mathcal{B}$ and provide the motivations for that choice (section \ref{sec:3_2_3_designing_b_for_grasping}).

%---------------------------------------%
% Driving B vs Describing B
%---------------------------------------%

\paragraph{Driving $\mathcal{B}$ vs Describing $\mathcal{B}$.}\mbox{}\\
\label{sec:3_3_1_1_driving_b_vs_describing_b}

In section \ref{sec:2_2_qd_hard_explore}, we presented the history of NS-QD methods and how the two families of methods resulted in different paradigms. Depending on the targeted overall goal, we can distinguish two main usages of the behavior space. When doing SOO or BSC, the behavior space guides the evolutionary process toward the optimal solutions or the exploration of an outcome space. In those cases, $\mathcal{B}$ plays a \textit{driving} role. Introduced in the seminal Novelty-Search paper (\cite{lehman2011abandoning}), this idea is still critical in recent works on NS-derivated methods (\cite{paolo2021sparse}). When doing RIBS, the behavior space depicts how diverse is a solution $\theta$ compared to $\theta'$ for the targeted task – through the comparison of $b_{\theta}$ and $b_{\theta'}$, therefore playing a \textit{describing} role. This idea can first be found in NSLC (\cite{lehman2011evolving}) and MAP-Elites (\cite{mouret2015illuminating}) papers, and is critical for all recent works on ME-derivated methods (\cite{fontaine2023covariance}, \cite{anne2023multi}). In brief, \textbf{a driving $\bm{\mathcal{B}}$ helps to discover solutions, while a describing $\bm{\mathcal{B}}$ allows to distinguish them.} In any case, $\mathcal{B}$ is always driving and describing, but algorithms almost always focus on one or the other usage of $\mathcal{B}$.

It is worth noting that some recent works proposed to explore multiple behavior spaces (\textit{multiBD}) to leverage the two usages of $\mathcal{B}$: \cite{kim2021exploration} explores both the last position of the ball thrown by the robot (driving $\mathcal{B}$) and the orientation of the end effector at the middle of the episode to generate diverse ways to throw the ball to a given position (describing $\mathcal{B}$). Similarly, NSMBS (\cite{morel2022automatic}) explores several behavior spaces to generate diverse grasps (describing $\mathcal{B}$) while also exploring the position of the object at the end of the episode to force the generation of successful grasp (driving $\mathcal{B}$). The multiBD paradigm raises many questions: How to explore several behavioral spaces efficiently? How to design or learn the most relevant driving or describing $\mathcal{B}$? More importantly, how to compare QD methods on the obtained results, and how to interpret the output of the algorithm?

This work studies sparse reward and interaction through the application of grasping in robotics. As the ultimate objective is to do RIBS, \textbf{a good describing $\mathcal{B}$ is required}. As grasping is a sparse reward task, \textbf{a good driving $\mathcal{B}$ is also required}. To let the abovementioned multiBD questions for future work, \textbf{the problem must be addressed through a single $\bm{\mathcal{B}}$ that is both driving and describing}.

%---------------------------------------%
% Task alignment
%---------------------------------------%

\paragraph{Task alignment}\mbox{}\\
\label{sec:3_3_1_2_task_alignment}

In QD methods, the behavior space $\mathcal{B}$ supports the exploration – either to push the solutions toward some part of the outcome space (driving $\mathcal{B}$) or to distinguish them (describing $\mathcal{B}$). Several works discussed the importance of \textbf{having a good driving $\bm{\mathcal{B}}$ for making those methods successful}. In particular, \cite{pugh2015confronting} shows that the success of QD methods requires that $\mathcal{B}$ must be “aligned with the notion of quality”. 

QD methods succeed on hard exploration problems because the exploration of  $\mathcal{B}$ guarantees to eventually find a successful solution. This matter has been discussed through the hypothesis of uniform sampling of $\mathcal{B}_{reach}$ by \cite{doncieux2019novelty}. We propose to merge the idea of alignment with the hypothesis of uniform sampling through the following definition:
\begin{definition}
  A behavior space $\mathcal{B}$ is aligned with a task submitted to a fitness function $f$, a success threshold $f_{s}$, and a goal space $\mathcal{G} = \left\{ b_\theta \in \mathcal{B} \mid f(\tau_\theta)>f_{s} \right\}$ if the probability $p_{\theta^{s}}$ to find a solution $\theta^{s}$ such that $b_{\theta^{s}} \in \mathcal{G}$ verifies $\displaystyle \lim_{n_e \to \infty} p_{\theta^{s}}=1$, where $n_e$ is the number of domain evaluations.
\end{definition}

Let us illustrate this idea with the experimental example given by \cite{doncieux2019novelty}: the two wheels navigation robot that has to reach a specific point of a given maze at the end of the episode. Let us assume the hypothesis of NS uniform coverage proposed in the paper. By taking $\phi_\mathcal{B}(\tau)=\left(x_T, y_T\right)$, we guarantee to eventually find a solution that is close enough to the goal point $g\in \mathcal{B}$ to verify the success criterion. However, let $\phi_\mathcal{B}(\tau)=\alpha_T$, $\alpha_i$ being the agent's orientation at time step $i$. In that case, we cannot guarantee that the agent will ever find a successful solution: we can generate diversity by rotating the robot at its initial position without moving from it.

Now, there are two major limitations to the above reasoning. Firstly, this hypothesis assumes that $\mathcal{G} \subset$ $\mathcal{B}_{reach}$, which is highly dependent on both the controller and the domain;
Secondly, Doncieux et al. hypothesis is about pure NS: no later study has extended this work to the overall QD paradigm, where quality optimization is usually orthogonal to pressure toward novelty. We here are interested in illuminating a behavior space. \textbf{Now that we consider sparse reward tasks, we want to make sure that the chosen $\bm{\mathcal{B}}$ can be illuminated in practice.}

The purpose of the present work is not to dig into those theoretical problems. What matters here is to stress the importance of the choice of $\mathcal{B}$: \textbf{by mostly working on navigation tasks, the QD literature in evolutionary robotics assumes that $\bm{\mathcal{G} \subseteq \mathcal{B}_{reach}}$ and make sure it is true in practice.} Note also that the Euclidean distance is a distance function that provides meaningful information on the exploration process: if the task is to reach a specific point at the end of the episode, the Euclidean distance on the last position allows accurate comparison of rollouts with respect to the considered task.

%---------------------------------------%
% Behavioral density
%---------------------------------------%

\paragraph{Behavioral density}\mbox{}\\
\label{sec:3_3_1_3_behavioral_density}

Similarly to fitness, some behavior functions do not always provide information the working algorithm can exploit. To address grasping, \cite{morel2022automatic} proposed several behavioral characterization – including the orientation of the robot's end effector when touching the object for the first time; if the object is not touched, the behavior descriptor is not defined. A behavioral characterization that always provides exploitable information can be described as follow:

\begin{definition}
  A behavioral function $\phi_{\mathcal{B}}$ is \textit{dense} if its support is equal to its domain (here if $\supp \phi_\mathcal{B} = \Theta$).
\end{definition}

Note that the navigation tasks usually addressed in QD all involve dense $\phi_{\mathcal{B}}$ (\cite{lehman2011evolving}, \cite{mace2023quality}, \cite{faldor2023map}, \cite{zardini2021seeking}). On the contrary, a behavioral function that does not always provide exploitable information can be defined as follow:

\begin{definition}
  A behavioral function that is not dense is called \textit{sparse}.
\end{definition}
In robotics, this setup can be referred to as a \textit{sparse interaction} problem, as sparse behavioral function corresponds to trajectories in which the robot did not interact with the entities of interest. Of course, \textbf{all sparse behavioral functions do not result in similar task difficulty}. To estimate the sparsity associated with $\phi_{\mathcal{B}}$, a ratio of evaluations that result in a behavioral signal can be computed (see \textit{outcome ratio} in section \ref{sec:4_3_2_metrics}).

%---------------------------------------%
% Behavioral completeness
%---------------------------------------%

\paragraph{Behavioral completeness}\mbox{}\\
\label{sec:3_3_1_4_behavioral_completeness}

\begin{definition}
  A behavioral characterization is \textit{complete} if the behavior space $\mathcal{B}$ is aligned with the targeted task and the behavioral function $\phi_{\mathcal{B}}$ is dense.
\end{definition}

All QD works that address sparse reward problems involve a complete behavioral characterization (\cite{lehman2011abandoning}, \cite{paolo2021sparse}, \cite{fontaine2023covariance}). In other words, the defined $\mathcal{B}$ ensures that \textbf{the algorithm can rely on an exploitable behavioral signal} throughout the evolutionary process, and that the exploration of $\mathcal{B}$ will \textbf{eventually result into the discovery of the solutions of interests} (e.g. the optimal solution, or all the solutions that validates a success criterion).

The present work argues that \textbf{a complete behavioral characterization cannot be trivially defined for many interesting tasks}. The next section elaborates on \textbf{why grasping is one of them}.

%Many problems that have not been addressed yet by the QD literature might not easily satisfy the above assumptions, making the exploration process easier and more interpretable, and in the case of NS,  guarantees that we will find a solution. \textbf{Grasping is a good example of a problem in which defining $\bm{\mathcal{B}}$ is challenging}; as discussed in the next section, we cannot easily define $\mathcal{B}$ to make it aligned with grasping.

%--------------------------------------------------------------------%
%               3.3.1) Defining B for grasping
%--------------------------------------------------------------------%

\subsubsection{Defining $\mathcal{B}$ for grasping:}
\label{sec:3_3_2_define_b_for_grasping}

The definition for $\mathcal{B}$ is crucial for making QD algorithms work well (\cite{pugh2015confronting}). The ideal $\mathcal{B}$ would be: 1) in \textbf{low dimension} (\cite{cully2022quality}), 2) \textbf{unique} (to avoid the complexity of multiBD), 3) \textbf{aligned with the task} (to efficiently drive the exploration), 4) \textbf{meaningful from the task perspective} (to generate a diversity of solution we are interested in), 5) and \textbf{easily interpretable}. This section elaborates on why misalignment cannot be avoided in grasping, and propose a behavior space that satisfies all the other mentioned criteria.

%---------------------------------------%
% Why misalignment cannot be avoided
%---------------------------------------%

\paragraph{Why misalignment cannot be avoided}\mbox{}\\

To address hard exploration problems, QD works on NS-derivated methods that rely on a behavior space $\mathcal{B}$ aligned with the targeted task. In these works, $\mathcal{B}$ is defined such that it can be inferred from the expression of the success criteria $f_c$. In \cite{paolo2021sparse}, all the success criteria depend on the last position of the object of interest (e.g. the end effector, a ball). $\mathcal{B}$ is thus aligned with the task: the exhaustive exploration of $\mathcal{B}$ defined as the agent's last position will eventually result in the discovery of the best-performing solutions.

A similar approach might be applied for grasping: infer from $f$ the trajectory's components that must be explored in order to find the best solutions. This work considers a grasping trajectory to be successful if a validation condition $\xi:\mathbb{R}^6\times\mathbb{R}^6 \rightarrow \left\{0, 1\right\}$  is verified for $N_g$ steps:
\begin{equation*}
\left ( \sum_{i=1}^{N_g} \mathbbm{1}_{\xi ( X_{obj}^{T-N_g+i}, X_a^{T-N_g+i}) } \right ) = N_g
\end{equation*}
where $X_{obj}^{i}$ and $X_{ee}^{i}$ are respectively the state of the object and the agent's end effector at step $i$. Both states are expressed as a 6 degrees-of-freedom pose, that is, the concatenation of a Cartesian position and its orientation in Euler angles. The grasping validation condition $\xi$ is verified if the end effector and the object are in contacts\footnote{Assuming the 3D models of the object and the robot are known, the contacts can be detected with the 6 degrees-of-freedom poses only.}, and if $z_{obj}^i>z_{obj}^0$, $z_{obj}^i$ being the $z$ component of the object's Cartesian position at step $i$. Inferring $\mathcal{B}$ from the above success criterion lead to:
\begin{equation*}
dim(\mathcal{B}) = N_g (dim(X_a) + dim(X_{obj}))
\end{equation*}
By setting $N_g=10$ steps, as used in this work experiments: 
\begin{equation*}
dim(\mathcal{B}) = 10 \times (6 + 6)= 120
\end{equation*}
The obtained value is way larger than low dimensional behavior spaces usually considered in QD, where $dim(\mathcal{B}) < 10$. QD methods are designed to operate on low dimensional spaces (\cite{cully2022quality}). Considering such a high dimensional $dim(\mathcal{B})$ raises many questions: Can QD methods correctly explore such a large space? Should the obtained behavioral vector be encoded in a lower dimension – falling into the problems of representation-learning for QD (\cite{cully2019autonomous}, \cite{paolo2020unsupervised})? Note that this $\mathcal{B}$ is designed to drive the exploration. Should a describing $\mathcal{B}$ also be defined – resulting in a multiBD problem?

%---------------------------------------%
% Proposed choice
%---------------------------------------%
\paragraph{Proposed choice}\mbox{}\\

The present work aims to study QD for grasping in a straightforward and interpretable manner. Therefore, \textbf{we have decided to set $\bm{\phi_\mathcal{B}(\tau)=X_{a}^{touch}}$, where $\bm{X_{a}^{touch}}$ is the position of the agent's end effector when touching the object for the first time in the episode}. This choice gives to $\mathcal{B}$ many of the expected properties: 1) it \textbf{keeps $\bm{\mathcal{B}}$ in low dimensions} ($dim(\mathcal{B})=3$, which is the same order of magnitude as standard QD problems); 2) it consists of a \textbf{single behavior space}; 3) it is a \textbf{good driving behavior}, as touching the object is a prerequisite for grasping – discarding from a behavioral perspective any trajectory that does not interact with it; 4) it is a \textbf{well-describing descriptor}, as the goal is to distinguish grasps from the position in which the end effector interacts with the object – allowing us to control the granularity of the generated diversity at a physically meaningful scale (e.g. $1\text{cm}^3$); and 5) it makes the \textbf{outcome easily interpretable and visualizable}.

%---------------------------------------%
% Consequences
%---------------------------------------%

\paragraph{Consequences}\mbox{}\\

A major drawback of the proposed behavior space is that for any given individual $\theta$, the resulting descriptor $b_\theta$ is not defined if the object is not touched throughout a trajectory $\tau_\theta$. But what makes the QD methods efficient on hard exploration problems is replacing a fitness signal with a behavioral one, in cases where the fitness landscape is deceptive or flat (\cite{lehman2011abandoning}). In cases where $\mathcal{B}$ is perfectly aligned with the task, the optimization process can be guided with novelty to explore a behavioral landscape that always provides exploitable information. \textbf{Taking $\bm{\phi_\mathcal{B}(\tau)=X_{a}^{touch}}$ makes the problem fall into a new kind of QD problem where the domain is submitted to a sparse behavior function (or sparse \textit{interaction})}. Another drawback is that the chosen \textbf{$\bm{\mathcal{B}}$ is misaligned with grasping}, as exhaustively exploring the space of first touching points on the object does not guarantee to find a successful grasp.

%Now that the motivation for the choice of $\mathcal{B}$ have detailed, we need to define the used fitness function, as well as the architecture of the policies $\pi_\theta$ that conditions agents' behaviors in the grasping environments. In the next section, we describe the open loop way-points controller used to define $\pi_\theta$.

\begin{figure}[t]
\begin{center}
\centerline{
 \includegraphics[width=\textwidth]{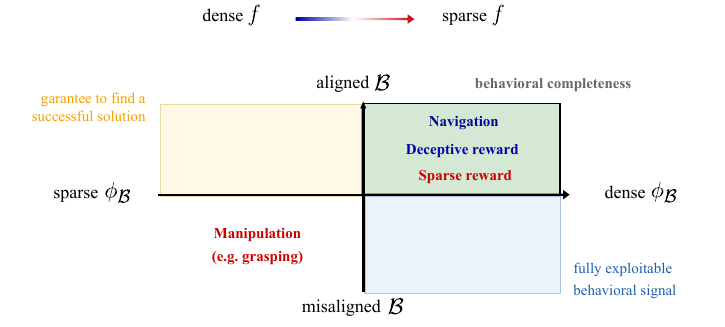}
}
\end{center}
\caption{\textbf{Challenges to address when using QD for grasping.} In standard QD benchmarks for evolutionary robotics, the behavioral characterization is complete: $\phi_\mathcal{B}$ is dense, and $\mathcal{B}$ is aligned with the task. This is true for rapid illumination of $\mathcal{B}$ in locomotion tasks (\cite{cully2022quality}), deceptive reward problems (\cite{lehman2011abandoning}), or sparse reward QD (\cite{paolo2021sparse}). By studying grasping, the present work argues that addressing QD with manipulation tasks can lead to sparse $\phi_\mathcal{B}$ and misaligned $\mathcal{B}$.}
\label{fig:opti_pb_diff_diag}
\end{figure}

Figure \ref{fig:opti_pb_diff_diag} gives an overview of the challenges to tackle when addressing grasping with QD. While standard benchmarks imply leveraging a complete behavioral characterization for generating diversity of solutions, or for solving sparse or deceptive reward tasks,\textbf{ grasping involves sparse $\bm{\phi_{\mathcal{B}}}$ and misaligned $\bm{\mathcal{B}}$}. Several questions arise: \textbf{can successful solutions be found with a misaligned $\bm{\mathcal{B}}$?} And \textbf{how to do rapid illumination of a behavior space if $\bm{\phi_{\mathcal{B}}}$ is sparse?} The next section describes the experimental protocol proposed to get empirical answers to those questions by studying QD methods performances on the task of grasping.

% À PARTIR DE LA :
% Faire le point avec un schéma récaptitulatif qui place la saisie par rapport aux pbs classiques, et les intérêts d'étudiers ce problème. 
% Mettre fitness et controlleur boucle ouverte en annexes. Donner le lien vers les annexes dans experiments/domains. Donner l'idée générale, et les quelques infos clefs. Et ça part sur les expés.

% Préciser les nejeux de cette étude expérimentale en fin de section "problème": étudier en pratique un cas où on a sparse reward et interaction.
% Comment les méthodes QD se comportent ?
% Qu'est-ce qui est le plus important pour bien fonctionner ?
% Y a-t-il des propriétés spécifiques à ces problèmes, que l'on n'a pas d'habitudes ?

% Fin visée :
% ME-scs marche le mieux (l'exploitation gourmandes de succès précédemment trouvés est très efficace)
% Malgré la non complétude de B, on a bien fait du RIBS (autrement dit, B est un bon driving et describing B)
% Propriété différente : les algos guidés pas la nouveautés ne sont pas les plus efficaces pour explorer un espace non-dense (nouveauté décevante)

%%%%%%%%%%%%%%%%%%%%%%%%%%%%%%%%%%%%%%%%%%%%%%%%%%%%%%%%%%%%%%%%%%%%%%%
%                           4) Experiments
%%%%%%%%%%%%%%%%%%%%%%%%%%%%%%%%%%%%%%%%%%%%%%%%%%%%%%%%%%%%%%%%%%%%%%%

\section{Experiments}
\label{sec:4_experiments}

This section describes the conducted experiments for evaluating how QD methods can address grasping. At first are described the grasping environments (section \ref{sec:4_1_environments}), then the compared methods (section \ref{sec:4_2_methods}), and finally the evaluation process (section \ref{sec:4_3_evaluation}).

%=====================================================================%
%                   4.1) Environments
%=====================================================================%

\begin{figure}[t]
\begin{center}
\centerline{
 \includegraphics[width=1.\textwidth]{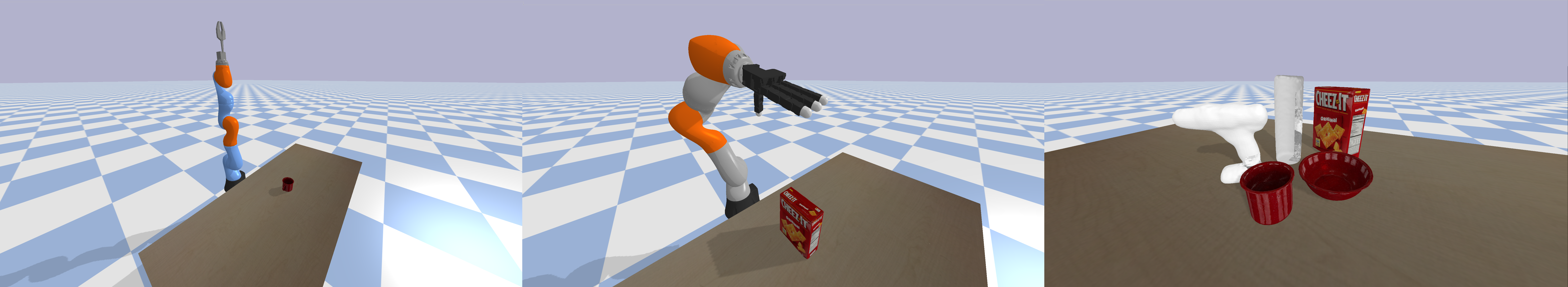}
}
\end{center}
\caption{\textbf{Studied domains.} (Left) The \textit{kuka\_wsg50\_far} environment consists of a parallel 1DoF gripper mounted on a kuka iiwa robotic manipulator. At the initial position, the end effector is far from the object. (Center) The \textit{kuka\_allegro\_close} robot environment consists of an Allegro 4-fingers dexterous hand mounted on a kuka iiwa. This end effector is initialized close to the object. (Right) The 3D models of the YCB objects (\cite{calli2015benchmarking}) to grasp: power drill, chips can, cracker box, mug, and bowl.}
\label{fig:studied_domains}
\end{figure}

\subsection{Environments}
\label{sec:4_1_environments}

The evaluated domains are grasping simulated scenes that rely on \textit{pybullet} (\cite{coumans2016pybullet}) (see Figure \ref{fig:studied_domains}). All scenes share a similar grasping scenario: a robotic arm is positioned close to a table with an object to grasp. Our study involves two \textit{kuka iiwa}, a 7-DoF robotic manipulator, with different end effectors: the first one is a parallel 2-fingers gripper (1-DoF); the other one is an Allegro hand, a 4-fingers dexterous robotics hand with 16-DoF (4-DoF per finger). Experiments have been made on 5 objects from the YCB-dataset (\cite{calli2015benchmarking}): \textit{chips can}, \textit{power drill}, \textit{mug}, \textit{bowl} and \textit{cracker box}. The first environment is initialized so the parallel gripper is far from the object (\textit{kuka\_wsg50\_far}). The dexterous hand is initialized closer to the object in the second environment (\textit{kuka\_allegro\_close}).

The grasping controller consists of an open-loop trajectory guided by 3 waypoints. The gripper position is initialized to open position. When the end effector first touches the object, the gripper is closed with constant force. This mechanism is inspired by the \textit{Palmar Grasp Reflex}, which makes newborn infants close their hands when pressure and touch are applied to the palm (\cite{futagi2012grasp}). \textbf{The controller is described in detail in the section \ref{sec:a1_architecture_of_policies_for_grasping} of the supplementary materials.} Finally, the fitness function consists of a normalized mixture of two sub-fitnesses that aim to minimize energy consumption and to minimize the variance of contact points between the end effector and the objects. Note that its value is set to $0$ if the object is not grasped. \textbf{The fitness function computation is detailed in section \ref{sec:a1_b_fitness} of the supplementary materials.}

%=====================================================================%
%                        4.2) Methods
%=====================================================================%

\subsection{Methods}
\label{sec:4_2_methods}

\begin{table}
\centering
\begin{tabular}{ ||c || c | c | c || c | c | c || c | c || c | c | c || }
\hline
\textbf{methods} & \multicolumn{3}{c||}{\textbf{root}}
            & \multicolumn{3}{c||}{\textbf{container}}
                    & \multicolumn{2}{c||}{\textbf{pop}}
                            & \multicolumn{3}{c||}{\textbf{goal}} \\
    \hline
   &  NS  & ME & \o &  
            UA & SA & \o & 
                PP & NPP & 
                RIBS & cvg($\mathcal{B}$) & $\theta^*$  \\
    \hline
Random   &  &   & $\bullet$  &   &  &  $\bullet$ &  $\bullet$  &  &   &   &  \\
    \hline
NS   & $\bullet$ &   &   &  $\bullet$ &  &  &  $\bullet$  &   &   & $\bullet$  & $\bullet$ \\
    \hline
Fit   &  &   & $\bullet$  &  $\bullet$  & &  & $\bullet$  &   &   &   & $\bullet$ \\
\hline
NSLC   & $\bullet$ &   &   & $\bullet$  & &  & $\bullet$  &   & $\bullet$  &   &  \\
\hline
NSMBS  & $\bullet$  &   &   & $\bullet$  & &  & $\bullet$  &   &   & $\bullet$  &  \\
\hline
SERENE   & $\bullet$  &   &   & $\bullet$  & &  & $\bullet$  &   & $\bullet$ &  &  \\
\hline
ME-rand  &   & $\bullet$  &   &   & $\bullet$  & &  & $\bullet$  & $\bullet$  &   &  \\
\hline
ME-scs   &   & $\bullet$  &   &   & $\bullet$  & &  & $\bullet$  & $\bullet$  &   &  \\
\hline
CMA-ES   &   &   & $\bullet$ &   &  &  $\bullet$ &  $\bullet$ &  &   &  & $\bullet$ \\
\hline
CMA-ME   &   &  $\bullet$ &   &   & $\bullet$  & &  & $\bullet$  & $\bullet$  &   &  \\
\hline
CMA-MAE   &   & $\bullet$  &   &   & &  &   & $\bullet$  & $\bullet$ &   &  \\
    \hline
\end{tabular}
\caption{\textbf{Taxonomy of the compared methods.} Each column corresponds to a family of QD method (see section \ref{sec:3_2_taxonomy_of_qd_methods}): the root algorithm (Novelty Search (NS), MAP-Elites (ME), or none of those two ($\o$)), the nature of the container (unstructured archive (UA), structured archive (SA), or no container ($\o$)), the population mechanism (population-based (PP) or directly sampled from the archive (NPP)), and the overall objective of the algorithm (rapid illumination of a behavior space (RIBS), dense coverage of a behavior space (cvg($\mathcal{B}$)), or finding the optimal solution ($\theta^*$).}
\label{table:cmp_method_taxonomy}
\end{table}

Table \ref{table:cmp_method_taxonomy} provides an overview of each of the studied methods with respect to the taxonomy proposed in section \ref{sec:3_2_taxonomy_of_qd_methods}. This table shows why NS and ME-derivated methods are usually called NS and QD-based methods, as most of those algorithms share common properties. However, we believe this matter is a chicken-and-egg problem: the lack of accurate distinction results in implicit design choices that do not question previously established algorithmic paradigms. 

This section provides details on the compared methods. All methods rely on the same behavioral characterization ($\phi_\mathcal{B}(\tau)=X_{a}^{touch}$). When not explicitly stated, the mutation is a Gaussian perturbation. We decided not to use the crossover to avoid adding too many complexities and hyperparameters that might affect the results. Some works in the field suggest that this could improve the performances (\cite{vassiliades2018discovering}); crossover is thus considered for future work. For comparison fairness, all NPP methods that can sample a variable number of individuals from the container are implemented such that this number matches the population size of PP methods.

\textit{\textbf{Random.}} Simple PP baseline that initializes its population randomly, evaluates it, sample offspring, evaluate them, and then randomly generate a new population for the next generation.

\textit{\textbf{NS.}} Similar to standard NS method (\cite{lehman2011abandoning}).

\textit{\textbf{Fit.}} Another common PP baseline that selects its population based on individuals' fitnesses (descending order). 

\textit{\textbf{ME-rand.}} Standard MAP-Elites algorithm (\cite{mouret2015illuminating}), that randomly samples individuals from the archive of elites.

\textit{\textbf{ME-scs.}} ME-rand does not seem suited to the targeted task, as the hard exploration nature of the problem will make the probability of sampling the best-performing individuals very low. To explore the potential of ME on grasping, a simple variant that selects in priority individuals $\theta_i$ such that $f(\theta_i)>0$ would be more suited to the targeted task. In practice, it randomly samples individuals from the successful ones in the container, and if there are not enough solutions to fill the "population" (set of individuals used to generate the offspring), it fills it with randomly sampled solutions – just like a standard ME-rand. This simple ME variant seems way more adapted to grasping, as the research will be strongly biased toward local regions around already successful solutions.

\textit{\textbf{NSLC.}} Similar to standard NSLC (\cite{lehman2011evolving}). Surprisingly, this seminal QD method has been left behind in favor of MAP-Elites (\cite{cully2022quality}). To our knowledge, no reference paper provides theoretical or experimental results that would justify dropping it when addressing a new QD problem. There are many algorithmic differences with ME: Pareto front selection over novelty and local quality, the way local quality is computed (how many neighbors have lower fitness than the evaluated individual), and its UA and PP backbone. This goal is not to compare those two approaches comprehensively but instead consider both methods as candidates for addressing grasping with QD. 

\textit{\textbf{NSMBS.}} NSMBS is the only QD method in the literature that is explicitly designed to address grasping (\cite{morel2022automatic}). It consists of an NS-derivated algorithm – PP and UA – that introduces two innovations: exploring multiple behavior spaces, and a specific selection operator that sequentially selects a behavior space and then makes a tournament-based novelty-guided selection. Note that an exhaustive comparison with QD methods is yet to be done. To get a fair comparison and easily interpretable results, NSMBS has been applied on a single behavior space ($\phi_\mathcal{B}(\tau)=X_{a}^{touch}$). NSMBS is thus similar to NS with a tournament-based selection. A comprehensive study on multiBD for grasping is considered for future work.

\textit{\textbf{SERENE.}} NS-derivated method specifically designed to address sparse reward problems (\cite{paolo2021sparse}). The method can be decomposed into two phases: an exploration phase consisting of a standard NS algorithm; and an exploitation phase, exploiting solutions found with non-null reward to initialize CMA-ES emitters to refine the solutions. To our knowledge, this is the only work specifically focusing on sparse reward context for QD. 

\textit{\textbf{CMA-ES.}} Covariance Matrix Adaptation Evolution Strategy (CMA-ES) is one of the best derivate-free optimization algorithms for continuous domains (\cite{hansen2016cma}). It models the sampling distribution of the population as a multivariate normal distribution, estimated from the previous generation's best-performing solutions. Even though this method is single objective-oriented, its impressive results on many problems made us use it as a baseline to estimate how good other methods are to explore, or to generate high-performing solutions. Note that CMA-ES is not a QD algorithm.

\textit{\textbf{CMA-ME.}} Covariance Matrix Adaptation MAP-Elites (CMA-ME) combines self-adaptation techniques of CMA-ES with diversity-maintaining techniques of MAP-Elites (\cite{fontaine2020covariance}). Despite the great results shown in the paper, recent work shows that this method struggle in sparse reward domains (\cite{paolo2021sparse}). This method is kept as a baseline to verify this weakness in new domains and emphasize key properties of other algorithms.

\textit{\textbf{CMA-MAE.}} Improved version of CMA-ME to address three of its limitations: quick abandonment of difficult-to-optimize objectives, inability to efficiently explore flat objective functions, and inefficiency on low-resolution archives. The results presented in this recently published paper (\cite{fontaine2023covariance}) makes CMA-MAE a state-of-the-art QD method. The fact that it has been explicitly designed to address flat fitness landscape scenarios makes it a serious candidate to tackle grasping. 

Given the good results of MAP-Elites, a study has also been conducted on the following variants: 

\textit{\textbf{ME-fit.}} Standard MAP-Elites that select solutions with respect to their fitness, sorted in descending order.

\textit{\textbf{ME-nov.}} Standard MAP-Elites that select solutions with respect to their novelty, sorted in descending order.

\textit{\textbf{ME-nov-fit.}} Standard MAP-Elites that sample solutions through a pareto-front non-dominated selection, using both novelty and fitness.

\textit{\textbf{ME-nov-scs.}} Standard MAP-Elites that select solutions with respect to their novelty, sorted in descending order – selecting successful solutions in priority (similarly to ME-scs).

As discussed in section \ref{sec:3_3_2_define_b_for_grasping}, $\phi_\mathcal{B}(\tau)=X_{a}^{touch}$ is defined for any sampled trajectory. The following design choice has thus been made: if a method requires a defined behavioral descriptor for an evaluated individual, its value is set to an arbitrary fixed value ($b_\theta = (0,0,0)$)). Otherwise, its value is left undefined, as it will be discarded through the upcoming behavioral instructions.

All methods have been implemented from scratch, except for the following ones: the official SERENE implementation has been used (\cite{paolo2021sparse}), as well as pyribs (\cite{tjanaka2023pyribs}) to get the official implementation of CMA-ME and CMA-MAE. Note that pyribs has also been leveraged for standard CMA-ES, relying on the theoretical equivalence of CMA-MAE with CMA-ES for $\alpha=0$ (\cite{fontaine2023covariance}). Details can be found in the publicly shared code.

To limit the energy consumption of this study, the experiments have been carried out on two evaluation budgets: 100k evaluations (long run) and 400k evaluations (very long run). The 100k evaluations experiments include the following methods: Random, NS, Fit, ME-rand, ME-scs, NSLC, NSMBS, SERENE, CMA-ES, CMA-ME, CMA-MAE. Methods tested on 400k evaluations depend on the best-performing methods obtained on 100k evaluations: ME-rand, ME-scs, ME-fit, ME-nov, ME-nov-fit, ME-nov-scs, CMA-ME, CMA-MAE.

%=====================================================================%
%                        4.3) Evaluation
%=====================================================================%

\subsection{Evaluation}
\label{sec:4_3_evaluation}

This section describes the evaluation protocol used in the experiments. At first is presented the framework proposed to distinguish the evaluation of QD algorithms from their internal components (section \ref{sec:4_3_1_algo_output}), then the computed metrics (section \ref{sec:4_3_2_metrics}), and finally the chosen hyperparameters (section \ref{sec:4_3_3_hps}). 

%--------------------------------------------------------------------%
%                        4.3.1) Algorithms output
%--------------------------------------------------------------------%

\begin{figure}[t]
\begin{center}
\centerline{
 \includegraphics[width=\textwidth]{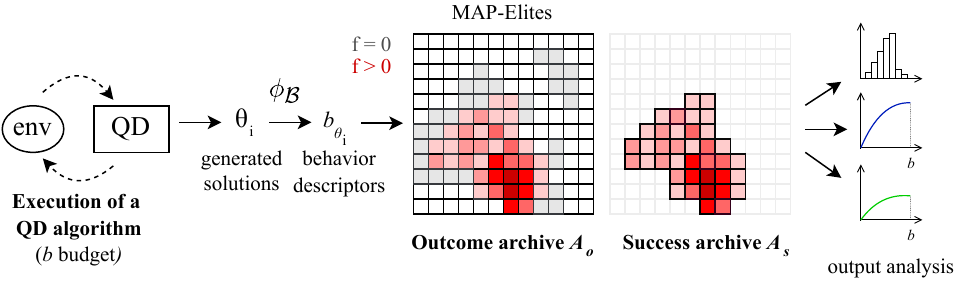}
}
\end{center}
\caption{\textbf{Algorithm evaluation framework.} To distinguish the algorithmic components from the evaluation container, all the generated solutions are considered as candidates to be added into an archive of elites (\cite{mouret2015illuminating}) here called the \textit{outcome archive}, after having projected them into the behavior space $\mathcal{B}$. Standard QD metrics can then be computed on this external archive. Grasping is a sparse reward task; what matters in this study is the set of successful solutions within $A_o$, called the \textit{success archive} $A_s$. This last archive is the ultimate output of the evaluated methods in the present work.}
\label{fig:algo_evaluation_framework}
\end{figure}

\subsubsection{Algorithm output}
\label{sec:4_3_1_algo_output}

As most QD works focus on either NS-derivated or ME-derivated methods, the running output is usually the container used by algorithms. RIBS-related works measure the structured archive coverage and qd-score (\cite{fontaine2023covariance}, \cite{mace2023quality}), while all works that focus on NS measure performances on the resulting unstructured archive or extracting from the evolutionary process the best-performing individual (\cite{paolo2021sparse}, \cite{conti2018improving}).

Comparing QD methods of different natures raises many issues: is the qd-score still a relevant metric when comparing SA and UA-based algorithms? How to compute $cvg(\mathcal{B})$? More generally, can the evaluation of performances be independent from the container used during the evolutionary process – discarding the constraint of operating in a fixed behavior space or even the requirement of using one? Recent works in the QD community stressed the rising need for distinguishing the output of the algorithms from the algorithmic modules themselves (\cite{fontaine2023covariance}).

In this work, \textbf{we introduce an evaluation procedure that distinguishes the algorithms' output from its internal components.} The evaluation framework is described in Figure \ref{fig:algo_evaluation_framework}, and basically consist of an external MAP-Elites that does not generate any new solution. The interaction between the environment and the QD algorithm is considered as a black box that generates solutions $\theta_i$ for a given budget $b$. All the solutions that have ever been generated during a given run are submitted to this evaluation procedure. Each obtained trajectory $\tau_{\theta_i}$ is then then projected into a behavior space $\mathcal{B}$, obtaining a behavior descriptor $b_{\theta_i}$. This descriptor is then considered for being added into an \textit{outcome archive} $A_o$, similarly to a standard MAP-Elites. The \textit{success archive} $A_s$ is then defined as : $A_s = \left\{ b_{\theta_i} \in A_o \mid f(\tau_{\theta_i})>0 \right\}$. Note that in the case of non-sparse reward domains, $A_s=A_o$.

%--------------------------------------------------------------------%
%                     4.3.2) Metrics
%--------------------------------------------------------------------%

\subsubsection{Metrics}
\label{sec:4_3_2_metrics}
Considering the above-proposed evaluation framework, the analysis is conducted on the following metrics:

\textit{\textbf{Coverage of the Outcome Archive.}} As the behavior space is here $\phi_\mathcal{B}(\tau)=X_{a}^{touch}$, computing the coverage of $A_o$ results here to answer the question: how diverse are the entry point for all the grasping attempts? The higher $cvg(A_o)$, the more first touching points have been discovered by the agent. 

\textit{\textbf{Coverage of the Success Archive.}} Computing $cvg(A_s)$ answers the following question: How many diverse successful grasps have been found? Interestingly, this metric appeared to be the most important one, as the qd-score for the chosen fitness function is aligned with this metric (see supplementary materials \ref{sec:a1_qd_scores}).

\textit{\textbf{Top-N fitnesses.}} To get an idea of how efficient the generated solutions are, the top-N fitnesses ever produced by each algorithm are also compared. This metric is an alternative to the qd-score to estimate the quality of the diverse generated solutions. Qd-score might be dominated by the number of found solutions, especially in difficult exploration tasks like grasping. It informs us on the number of successful grasps found, but not on their performances.

\textit{\textbf{Environment difficulty.}} To evaluate the challenges associated with each studied task, the environment difficulty is evaluated through two metrics: the \textit{outcome ratio} $\eta_o$ and the \textit{success ratio} $\eta_s$. To compute those metrics, $N_{ed}$ random individuals $\theta_i$ are generated by sampling from a uniform distribution within the genotype space $\Theta$. The obtained $\theta_i$ are then evaluated on the environment $\mathcal{M}$, getting the resulting number of individuals that successfully touched the object ($n_o$) and the number that grasped it ($n_s$). The ratios are then computed as follows:

\begin{equation*}
    \eta_o = \frac{n_o}{N_{ed}} \hspace{1cm} \eta_s = \frac{n_s}{N_{ed}} 
\end{equation*}

Those metrics stress whether an environment is submitted to sparse behavior function ($\eta_o \rightarrow 0$) or not ($\eta_o \rightarrow 1$). Similarly, $\eta_s$ expresses how sparse the problem is: the closer to $0$ the sparser, the closer to $1$ the denser.

%--------------------------------------------------------------------%
%                     4.3.3) Hyperparameters
%--------------------------------------------------------------------%

\subsubsection{Hyperparameters}
\label{sec:4_3_3_hps}

\textit{\textbf{QD methods.}}
Notations for hyperparameters are the following: $\mu$ is the population size, $\lambda$ the number of offspring, $n_A$ the number of individuals added to the archive at each generation, $k$ the number of neighbors considered for novelty computation, $N_{rt}$ the maximum number of rollouts. We set: $\mu=\lambda=100$, $n_A=40$, $k=15$. All offspring are mutated with a probability $ind_{pb}=0.3$ to modify each gene. For a fair comparison, all ME-derivated methods sample $\mu=\lambda$ individuals for offspring generation at each iteration. The mutation operator applied by default to all the methods is a Gaussian perturbation of $0$ mean and $0.5$ standard deviation\footnote{This standard variation has been chosen after doing a grid search on tested methods, selecting the value that maximizes $cvg(A_s)$ on the maximum number of methods avec 20k evaluations. It is worth noting that a too-small std prevents exploration, while a too-large std constrains the waypoints at the edge of the operational space.}. The archive size is unbounded for UA-based methods. For NSMBS, the tournament size for selection is set to $15$. For NSLC, we use $50$ neighbors to estimate local quality due to the sparsity of the task. The same parameters as SERENE paper have been used (\cite{paolo2021sparse}): the chunk size is set to $1000$, the emitter population len to $6$, and the same $k$ as other methods for estimating novelty. $5$ individuals are added to the archive at each iteration. For CMA\_* variants, we used the same parameters as in the papers (\cite{fontaine2020covariance}, \cite{fontaine2023covariance}): The emitter batch size is set to $36$, and the number of emitters to $15$. For CMA\_MAE, $f_{min} = -1$ and $\alpha=0.01$.

\textit{\textbf{Grasping domains.}}
The boundaries of the structured archive match the operational space. To get a precision of $1$cm³ for contact points, the number of bins per dimension is ($n\_bins_x$, $n\_bins_y$, $n\_bins_z$) = (24, 25, 25). Parameters used for normalizing energy consumption fitness for each robot are provided in the shared code on Github. The operational space is set as a box of (dx, dy, dz) = (1, 0.7, 0.5) meters on the top of the table. The object is initialized at the center. Robots are controlled in position such that their cartesian target cannot be set out of a virtual box within the operational space: (dx, dy, dz) = (0.8, 0.4, 0.4) meters for Allegro, (dx, dy, dz) = (1, 0.7, 0.5) for the 2-fingers grip. For the dexterous allegro hand, 6 grasping primitives have been defined: 1) index and thumb closure; 2) middle finger and thumb closure; 3) thumb and last finger closure; 4) thumb, index, and mid; 5) thumb; mid and last; and 6) all fingers closure. The episode length is $T=2000$ for \textit{kuka\_wsg50\_far} and $T=1500$ for \textit{kuka\_allegro\_close}.

\textit{\textbf{Environment difficulty.}} 400k randomly sampled trajectories have been deployed on all the considered domains. The controller and parameter space are the same as those used for the QD evaluation methods.

\textit{\textbf{Outcome archive.}} The archive sampling matches the one used on ME variants described above.

%%%%%%%%%%%%%%%%%%%%%%%%%%%%%%%%%%%%%%%%%%%%%%%%%%%%%%%%%%%%%%%%%%%%%%%
%                           5) Results
%%%%%%%%%%%%%%%%%%%%%%%%%%%%%%%%%%%%%%%%%%%%%%%%%%%%%%%%%%%%%%%%%%%%%%%

\section{Results}
\label{sec:5_results}

This section presents the obtained results. Here is what can be expected regarding the literature:
\begin{itemize}
    \item NSMBS, CMA-MAE, and SERENE are the most promising methods to do RIBS on grasping; they should dominate other methods on the evaluated metrics;
    \item NS should perform poorly on $cvg(A_s)$, as NSMBS paper shows that NS struggles to generate many successful solutions on grasping (\cite{morel2022automatic}). However, it should lead to a high value of $cvg(A_o)$, as NS is directly optimizing the exploration of $\mathcal{B}$ (\cite{paolo2021sparse}).  
\end{itemize}

The section \ref{sec:5_1_environment_difficulty} presents the result on the difficulties associated with each domain. The results obtained on 100k evaluations are provided in section \ref{sec:5_2_comparison_sota_qd_methods}. The results obtained on 400k evaluations are provided in section \ref{sec:5_3_impact_of_the_selection_operator}.

%=====================================================================%
%                   5.1) Environment difficulty
%=====================================================================%

\subsection{Environment difficulty}
\label{sec:5_1_environment_difficulty}

\begin{figure} %[t]
\begin{center}
\centerline{
 \includegraphics[width=0.6\textwidth]{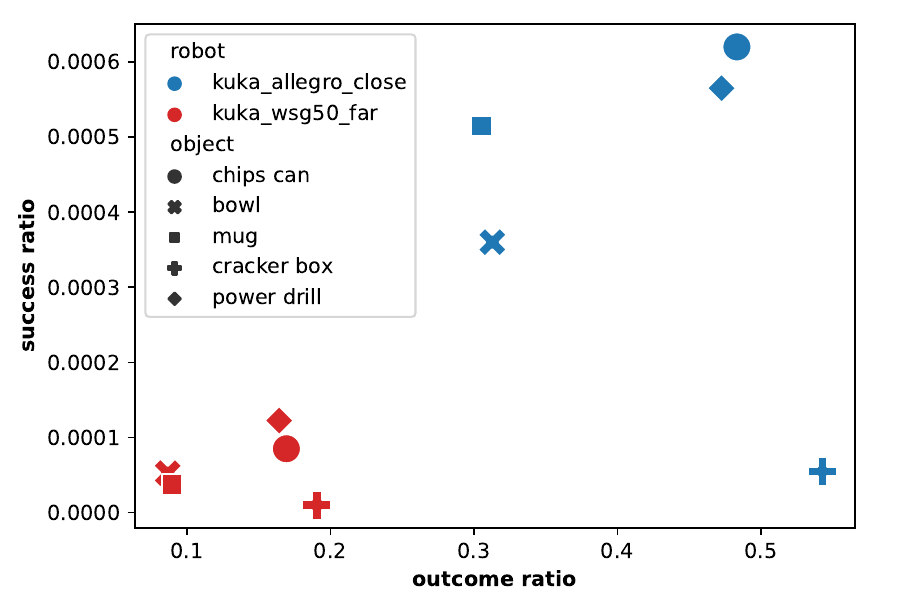}
}
\end{center}
\caption{\textbf{How challenging are the studied tasks.} We indeed are in a sparse reward context, as a few of the sampled trajectories lead to a non-null fitness. Plus, many deployed trajectories do not provide any behavioral information to exploit, making the exploration even more difficult. The higher the outcome ratio, the higher the success (Pearson correlation $r=0.61$ with $p<0.06$). Overall, \textit{kuka\_allegro\_close} domain is easier than \textit{kuka\_wsg50\_far} – except for the Allegro on the cracker box, in which the object is easy to touch but hard to grasp.}
\label{fig:env_difficulty_scatter}
\end{figure}

Figure \ref{fig:env_difficulty_scatter} gives an overview of the challenges associated with each domain. The outcome ratio on \textit{kuka\_allegro\_close} is significantly higher than those obtained on the \textit{kuka\_wsg50\_far}: between 31\% and 54\% of the randomly sampled trajectories touches the object on the first domain, while between 9\% and 19\% on the other one. In both cases, \textbf{many trajectories do not provide a behavioral signal that the algorithm can exploit to guide the exploration toward high-performing solutions}. As expected, the smaller objects (bowl and mug) lead to lower $\eta_o$ than the larger ones (power drill, cracker box, and chips can). \textbf{All domains are submitted to sparse rewards}, as the higher value of $\eta_s$ is close to 0.06\% of success, meaning that in a pure random search context, about 99.94\% of trajectories result in a null fitness.

Overall, the domains in which the object is easier to touch are those in which the object is easier to grasp (Pearson correlation between $\eta_o$ and $\eta_s$: $r=0.61$ with $p<0.06$). The Allegro hand on the cracker box is a remarkable outlier, though: it is easier to touch it ($\eta_o=0.54$) than any other objects on that gripper, but it is about 10 times more difficult to grasp it than on any other tested objects. The easier domains are (\textit{kuka\_allegro\_close}, chips can) and (\textit{kuka\_allegro\_close}, power drill), with respectively 48\% and 47\% of chance to randomly touch the object, and about 0.062\% and 0.057\% of chance to grasp it. (\textit{kuka\_allegro\_close}, bowl) can be seen as a intermediate-difficulty environment ($\eta_o=0.31$ and $\eta_s=0.00036$). The most challenging domains are the \textit{kuka\_wsg50\_far}-based setups, especially on the cracker box on which the success ratio reaches its smallest value ($\eta_s=0.00001$). The most challenging environment regarding the behavioral sparsity are (\textit{kuka\_wsg50\_far}, mug) and (\textit{kuka\_wsg50\_far}, bowl). It is also worth noting that a similar difficulty pattern can be seen on both robots for the studied object: the chips can and the power drill are the easier objects to address; the cracker box and the mug are more challenging tasks regarding both behavioral and fitness sparsity; and the cracker box is the most difficult to grasp object – despite of its high probability to touch it. The ratios obtained on the cracker box can be explained by the fact that there is a high probability of making the object fall, resulting in an impossible-to-grasp state for those grippers – while the chips can and the power drill are less challenging as they can still be grasped after having made them fell on the table.

Those metrics provide insight into key properties of the studied environments: \textbf{grasping is indeed submitted to sparse reward}. Plus, \textbf{the chosen behavior space} (see section \ref{sec:3_3_2_define_b_for_grasping}) \textbf{results in a flat behavioral landscape} too – as the object is not always touched. It actually mirrors the nature of the task and its inherent challenges, as the reach-and-grasp sequence is way more complex than adults think it is: it takes about 4 months for a baby to be able to reliably deploy this skill (\cite{needham2023babies}).

The next section presents the result obtained after 100k evaluations on the compared methods.

%=====================================================================%
%           5.2) Comparison of state-of-the-art QD methods
%=====================================================================%

\begin{figure} %[t]
\begin{center}
\centerline{
 \includegraphics[width=1.0\textwidth]{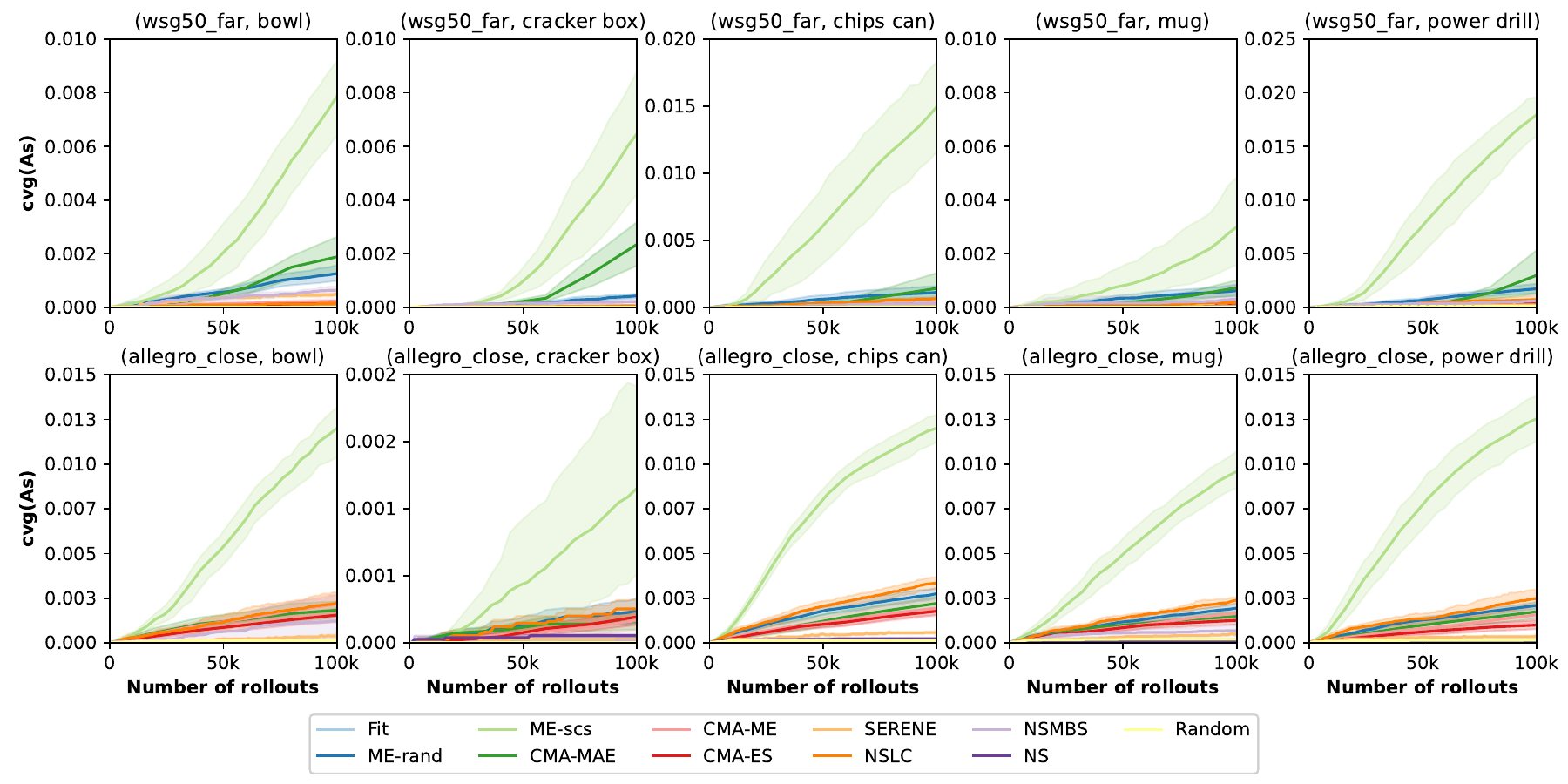}
}
\end{center}
\caption{\textbf{Coverage of the success archive throughout the evolutionary process.} Over 10 seeds. ME-scs dominate all the compared methods by a large margin, even the state-of-the-art QD methods designed to address sparse reward domains (CMA-MAE, SERENE, NSMBS).}
\label{fig:exp1_cvg_As}
\end{figure}

\begin{figure} %[t]
\begin{center}
\centerline{
 \includegraphics[width=1.0\textwidth]{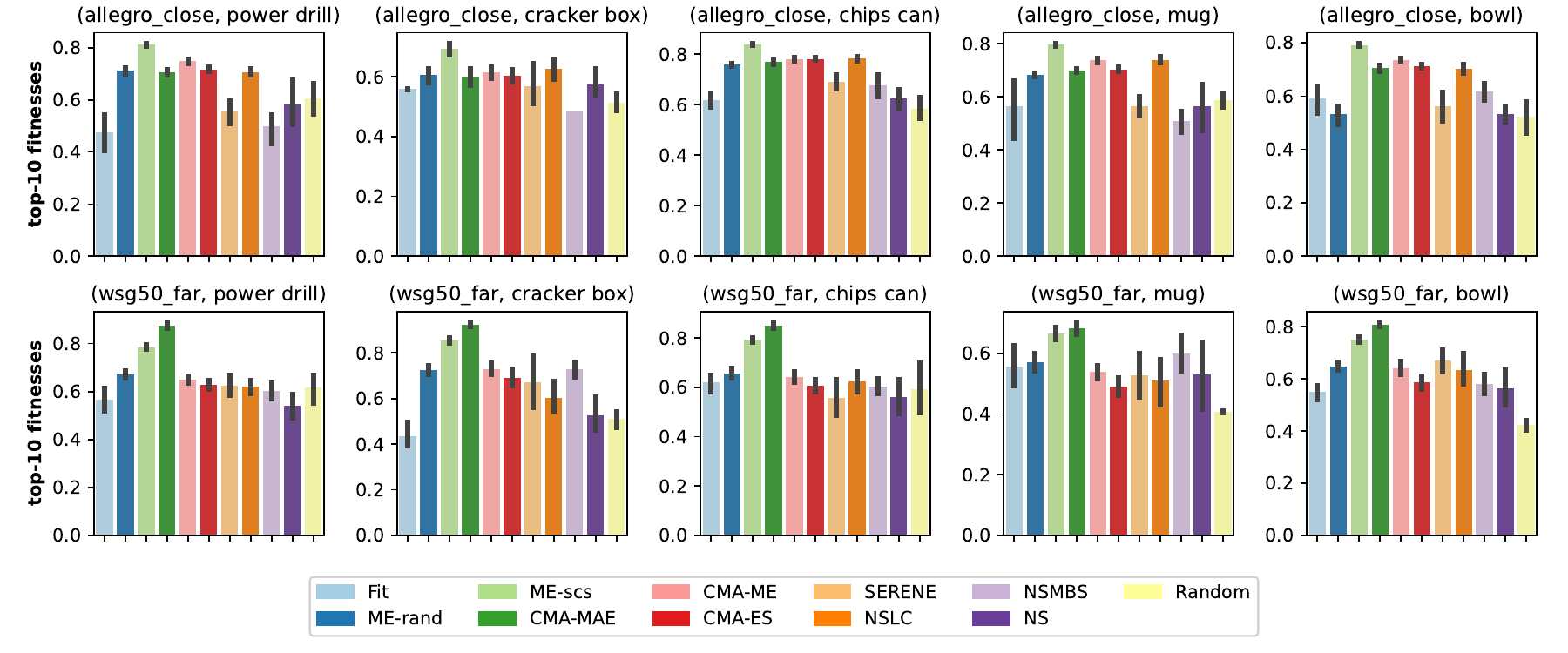}
}
\end{center}
\caption{\textbf{Fitnesses of the top-10 best performing individuals obtained after 100k evaluations.} ME-scs and CMA-MAE outperforms other methods after this number of evaluation, obtaining high-performing solutions with comparable fitnesses.}
\label{fig:exp1_top10_fit}
\end{figure}

\begin{figure} %[t]
\begin{center}
\centerline{
 \includegraphics[width=0.9\textwidth]{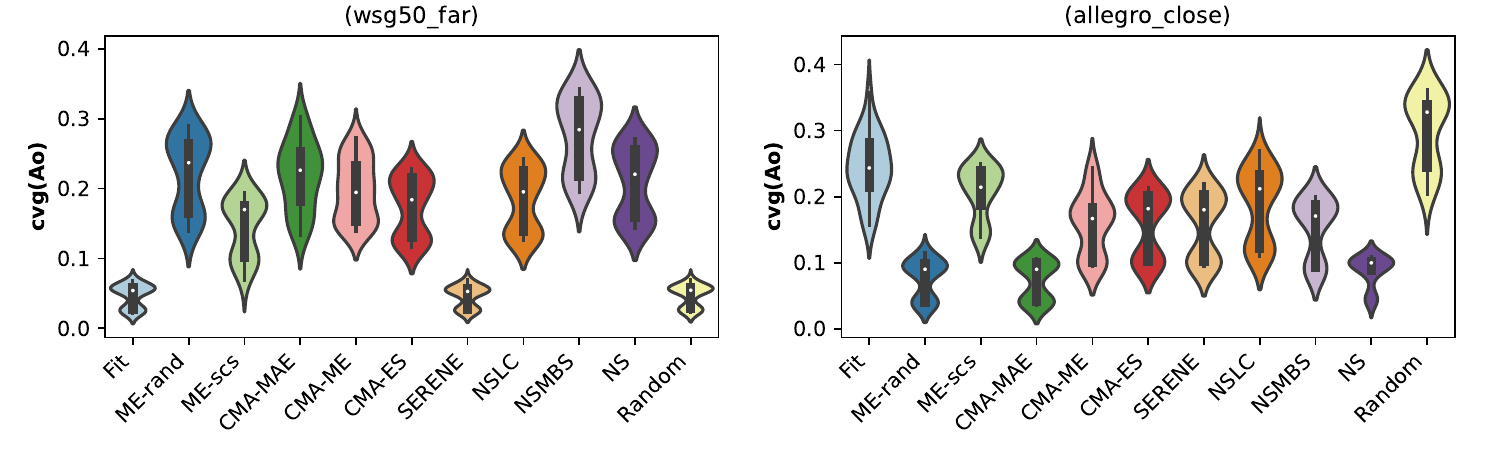}
}
\end{center}
\caption{\textbf{How do QD methods explore the objects surface.} Coverage of the output archive after 100k evaluations, over 10 seeds. Even though ME-scs crushes other methods on the generation of successful grasps, its exploration of $\mathcal{B}$ is not significantly higher than other methods – meaning that ME-scs prioritize parents that are the more likely to mutate into successful individuals.}
\label{fig:exp1_cvg_Ao_last_eval}
\end{figure}

\begin{figure} %[t]
\begin{center}
\centerline{
 \includegraphics[width=1.0\textwidth]{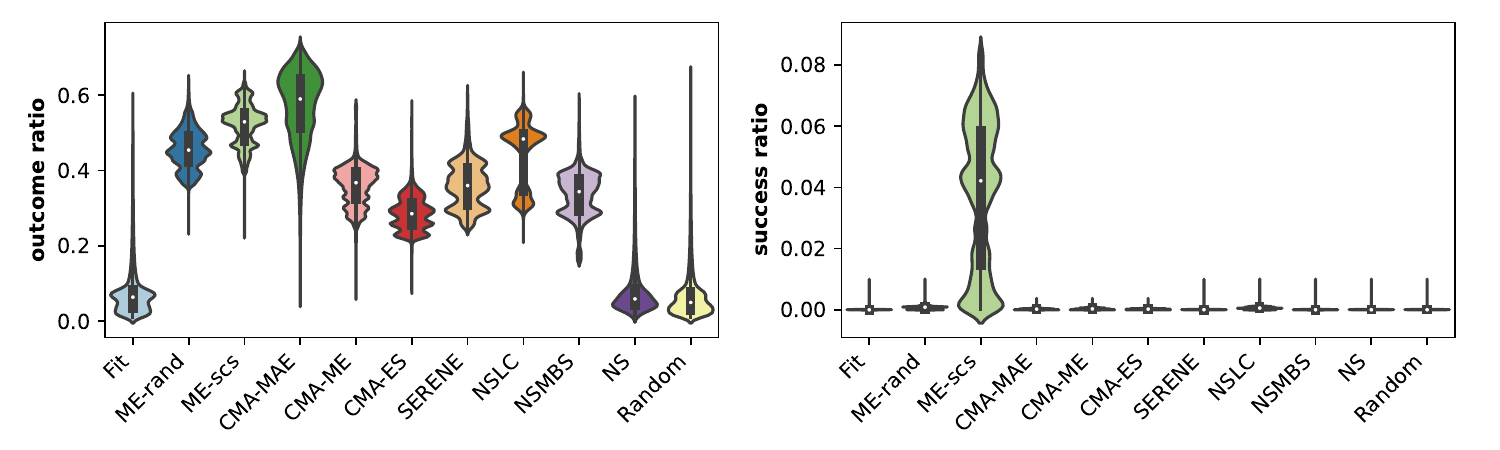}
}
\end{center}
\caption{\textbf{Distribution of success ratios and outcome ratios for each method on \textit{kuka\_allegro\_close}.} Aggregated over all objects throughout 100k evaluations over 10 seeds. By prioritizing successful solutions to generate offspring, ME-scs favors having a high probability of success. NS pushes its population close to the non-eligible regions of the search space. Quality-oriented emitters constrain the exploration in promising regions. CMA-MAE explores efficiently $\mathcal{B}$ but does not generate as many successes as ME-scs on those domains.}
\label{fig:exp1_outcome_succes_ratio_allegro}
\end{figure}

\subsection{Comparison of state-of-the-art QD methods}
\label{sec:5_2_comparison_sota_qd_methods}

\textit{\textbf{Generation of diverse successful solutions.}} In Figure \ref{fig:exp1_cvg_As} are the obtained coverage of the success archive for each method throughout the evolutionary process. The most striking result is that \textbf{ME-scs is dominating all the compared methods on all the evaluated domains by a large margin}. CMA-MAE outperforms all other methods except for ME-rand, which reaches similar coverage for several environments. SERENE is way below, and so is NSMBS. Another interesting point is that NSLC, the NS-derivated RIBS method, does not succeed as well as the ME-derivated variant. We attribute this to its selection process and its PP nature. On such a hard exploration problem, it is difficult to reach non-null fitness – such that the local quality score (\cite{lehman2011evolving}) is likely to either be at the maximal value (i.e. the individual is dominating all its neighbors) or to the lower value (i.e. the individual has a fitness lower or equal to all its neighbors). Consequently, the selection is always guided by novelty, making NSLC acts similarly to a standard NS. Note that the PP nature of both algorithms is prone to forgetting, as lower quality solutions that are not likely to generate new successes might be preferred to solutions with a higher quality and/or that are more likely to produce new successes because of the local quality issue. Lastly, NSLC optimizes the novelty and the quality of its current population. In contrast, ME directly optimizes a container with the same structure as the outcome archive to illuminate. 

\textit{\textbf{Generation of high-performing solutions.}} Figure \ref{fig:exp1_top10_fit} provides the distribution of top-10 fitnesses obtained with the evaluated methods. Again, \textbf{ME-scs outperforms all other methods}. Even CMA-MAE generated less performing solutions, showing that \textbf{the ME-scs ability to produce a large success archive lead to the emergence of high performing solutions}. The simple fitness-guided evolutionary algorithm (Fit) did not perform better than a random process, which is expected in hard exploration domains (\cite{lehman2011abandoning}). What is more unexpected is that NS did not perform any better than those two, while this method is meant to find high-quality solutions in hard exploration problems. We attribute this performance to the choice of $\mathcal{B}$. \textbf{As $\bm{\mathcal{B}}$ is here not aligned with the task} – there exist solutions that touch the object from any entry point without resulting into a grasp – \textbf{there is no guarantee that NS will eventually find the best-performing solutions.} 

Another interesting result is the poor performance of NSMBS. It is worth noting that NSMBS slightly outperforms NS in all the domains. The unique difference between those two methods here is the selection operator: the tournament selection of NSMBS by-pass the issue of non-eligible solutions – that must have a valid behavior descriptor for NS to compute each individual's novelty. The arbitrarily set value for non-eligible solutions creates a strong bias of novelty within the archive, pushing the exploration away from those values. On the contrary, NSMBS discards the non-eligible solutions from selection, as the tournament's candidate individuals are sampled from eligible individuals only (\cite{morel2022automatic}). It forces the exploration process toward eligible solutions without biasing the estimation of novelty within $\mathcal{B}$. Nevertheless, NSMBS performances are far from the best-performing ones. Those results discard NSMBS from the most promising QD methods to address grasping, stressing what should be its main interest: the multiBD context (see discussion in section \ref{sec:6_1_4_nsmbs}). 

SERENE reached slightly better performance than NS and got similar results to NSMBS. It matches what has already been discussed in the original paper: SERENE struggles to refine solutions in hard exploration domains (\cite{paolo2021sparse}). It initially behaves just like a standard NS. As soon as non-null fitness solutions have been found, they are evaluated as candidates for budget assignment to do local optimization. The problem is that in such a hard exploration domain as grasping, a successful solution might not lead to any successful one in a few 1-mutation trials. We attribute the obtained results to this phenomenon: in most cases where NS found successes, the candidates are evaluated as non-promising solutions for local optimization, making the method fall back to a standard NS. It sometimes allows local optimization, which explains the slight performance gain compared to NS on all the domains.

\textit{\textbf{Exploration of the object surface.}} Figure \ref{fig:exp1_cvg_Ao_last_eval} shows the coverage of the output archive after 100k evaluations. Interestingly, ME-scs does not dominate other methods on that metric. It shows that \textbf{by selecting successful solutions from the archive, ME-scs focuses on the regions of the object surface which are more likely to result in successful grasps}.

Surprisingly, methods derivated from NS do not explore the object surface better than other methods. On \textit{kuka\_wsg50\_far}, NS does not report a higher coverage than most quality-guided methods. Even more surprising is that \textbf{NS reports one of the lowest $\bm{cvg(A_o)}$ on \textit{kuka\_allegro\_close}.} Pure NS or NS variants that are not often submitted to a non-null quality signal are however optimizing $cvg(\mathcal{B})$ directly. NS-derivated methods which are also guided by quality have generated a significant number of successful solutions: SERENE, NSLC, and NSMBS dominate NS on $cvg(A_s)$ and on the top-10 fitnesses on many domains. They have therefore been confronted to non-null fitnesses, limiting their exploratory capabilities. But no straightforward answer can be brought to NS performances. Results given on the outcome ratio provide hints to explain those results.

\textit{\textbf{Outcome and success ratios.}} In Figure \ref{fig:exp1_outcome_succes_ratio_allegro} are displayed the outcome and success ratios measures for each algorithm throughout the evolutionary process on the \textit{kuka\_allegro\_close} domains. The selection operator of ME-scs leads to a probability of generating success $\eta_s = 3.7\%$. ME-scs is above 60 times more sample efficient than random sampling in $\Theta$ on the easier domain and 50 times more sample efficient than the second higher success ratio on that domain (ME-rand). But there is no statistically significant difference between NS and Random on the outcome ratio. Figure \ref{fig:exp1_cvg_Ao_last_eval} shows that NS does not behave as a Random search on this problem, the reported $cvg(A_o)$ is different for both methods. It means that \textbf{NS pressure for novelty has a deteriorating impact on $\bm{cvg(A_o)}$ in this context} – leading to an exploration of $\mathcal{B}$ that is similar to quality-guided methods (\textit{kuka\_wsg50\_far}) or significantly worse (\textit{kuka\_allegro\_close}). Such a result goes to the opposite of the strong exploratory power of NS that has been demonstrated in literature so far (\cite{paolo2021sparse}, \cite{doncieux2019novelty}). \textbf{We attribute the deteriorating impact of NS to its pressure for evolvability.} This point is discussed in section \ref{sec:6_2_detrimental_role_of_novelty_sparse_domains}.

The resulting QD-scores are provided in supplementary materials (Figure \ref{fig:a1_exp1_qd_score}). In this case, this metric is dominated by the number of found successful solutions, preventing us from relying on this measure to evaluate the ability of the tested methods to generate a set of diverse and high-performing solutions. Thus, the results only reflect the size of $A_s$, which is redundant with the coverage information.

The leading performances obtained with ME-scs invite us to investigate the role of the selection operator on the illumination of $A_s$ through ME variants. This second experiment compares several ME variants on longer runs (400k evaluations). Are also included the second-best performing solution (CMA-MAE) and the anterior version of this method (CMA-ME) to get more clues on how those state-of-the-art approaches behave in those domains.

%=====================================================================%
%            5.3) Impact of the selection operator QD
%=====================================================================%

\begin{figure} %[t]
\begin{center}
\centerline{
 \includegraphics[width=1.0\textwidth]{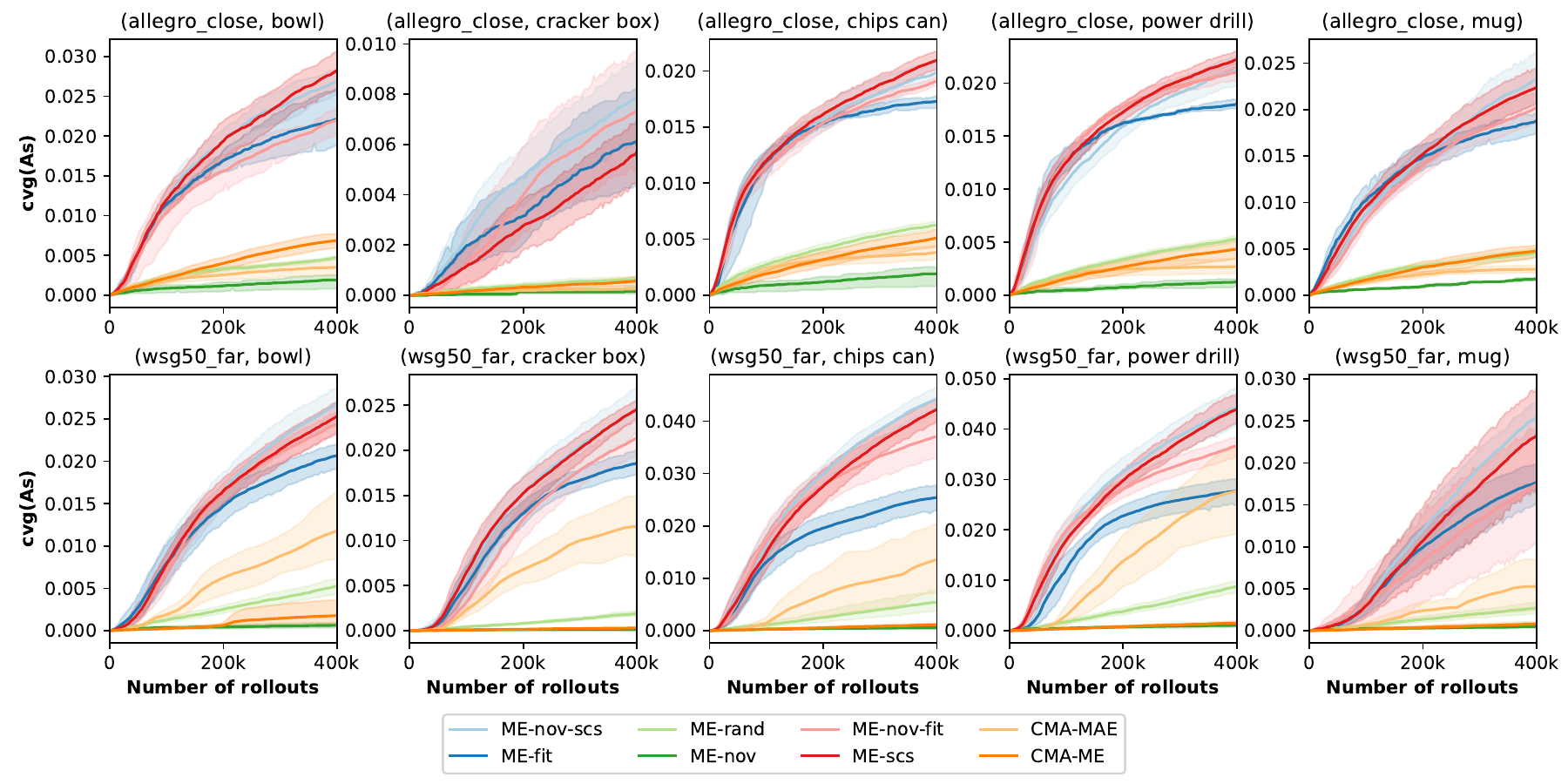}
}
\end{center}
\caption{\textbf{Impact of the selection operator on ME's ability to generate diverse, successful solutions.} Over 10 seeds. All ME variants that select non-null fitness solution in priority outperforms other methods. Prioritizing fitness can stick the exploration into local minima while selecting successful solutions regardless of their performance increases the success archive's size continuously.}
\label{fig:exp2_cvg_As}
\end{figure}

\begin{figure} %[t]
\begin{center}
\centerline{
 \includegraphics[width=1.0\textwidth]{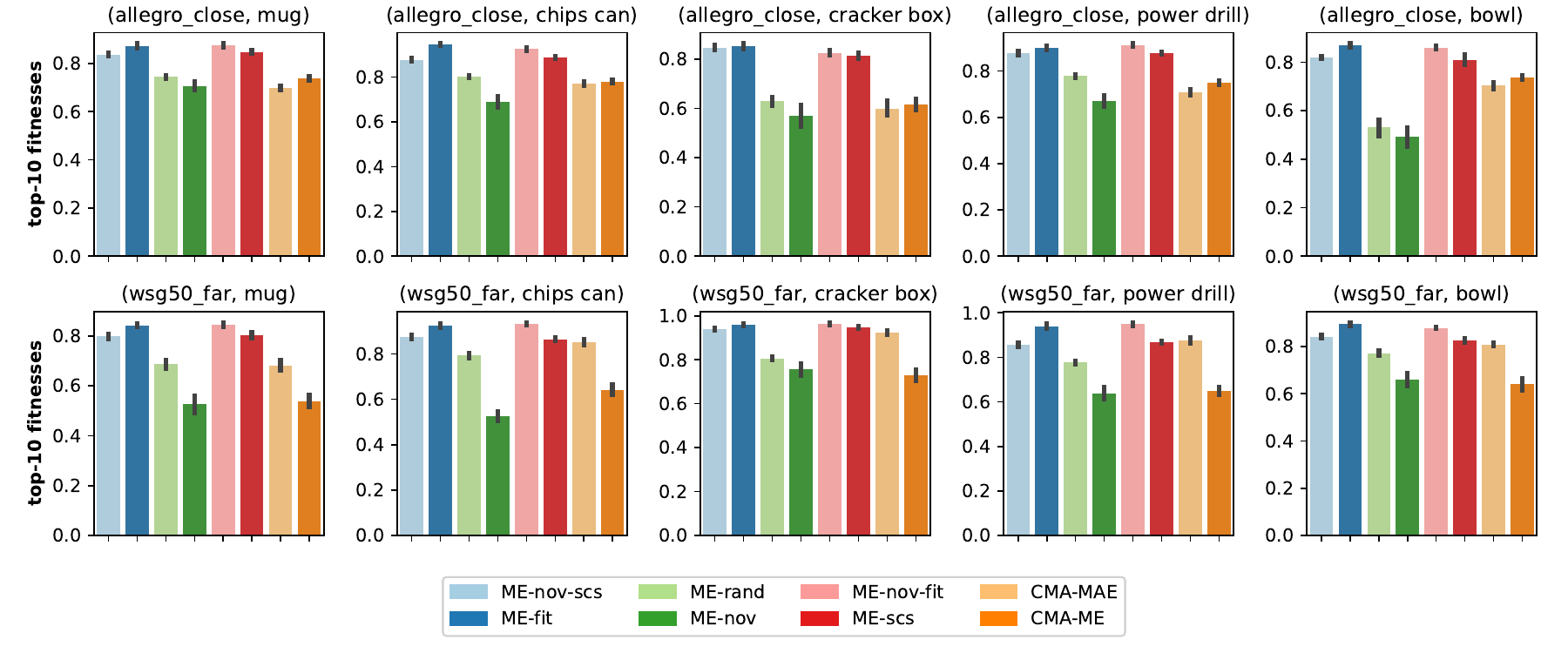}
}
\end{center}
\caption{\textbf{Impact of the selection operator on high performing solutions fitnesses.} Over 10 seeds, after 400k evaluations. ME-scs outperforms CMA-MAE on longer runs. ME-fit-based variants generate slightly better solutions than ME-scs-based ones.}
\label{fig:exp2_top10_fit}
\end{figure}

\subsection{Impact of the selection operator}
\label{sec:5_3_impact_of_the_selection_operator}

\textit{\textbf{Generation of diverse successful solutions.}} Figure \ref{fig:exp2_cvg_As} shows the coverage of the success archive. \textbf{Variants that select in priority the successful solutions (ME-scs, ME-nov-scs) dominate all other methods.} Prioritizing the most novel solution among the successful ones does not lead to a statistically significant difference. Selecting individuals with respect to their fitness (ME-fit and ME-nov-fit) leads to better results than other variants, except for the success-based ones. By selecting the best-performing individuals among the already discovered ones, the evolutionary process can get stuck in local minima – preventing the generation of new successful solutions. 

\textit{\textbf{Generation of high-performing solutions.}} Figure \ref{fig:exp2_top10_fit} shows the obtained distribution of top-10 fitnesses. It can be seen that \textbf{ME variants that prioritize fitness lead to the generation of better-performing individuals than success-based selection.} The two ME-scs variants reached comparable results. There are no statistically significant differences between ME-fit variants too. This result is unsurprising: the pressure for novelty is applied to already successful solutions for those 4 variants\footnote{A multi-objective solution that balances the pressure on performing/successful individuals with the pressure for novelty on non-successful individuals could have been set. As the scope of this work is rapid illumination of a success archive for grasping, the results obtained with novelty-guided solutions are not prone to stimulate too much effort in that research direction.}. 

The \textbf{poor performances of ME-nov on both $\bm{cvg(A_s)}$ and top-10 fitnesses enforce the result obtained with NS}: pressure for novelty might lead to deteriorated exploration performances. Interestingly, a ME-derivated method shares similar properties with an NS-derivated one. There are differences between ME-nov and NS beyond the nature of the archive: individuals compete on fitness in their behavioral niche, while NS does not. Discussion on the impact of novelty-based selection under in flat behavior landscape is provided in section \ref{sec:6_2_detrimental_role_of_novelty_sparse_domains}.

CMA-MAE is outperformed by all success or fitness-guided ME variants. It is worth noting that CMA-MAE consistently generates better high-performing solutions than CMA-ME on the most challenging robot (\textit{kuka\_wsg50\_far}). However, the opposite can be seen on the other robot. The method that dominates this other for generating high-performing solutions is also the one that reached the higher coverage of the success archive (see Figure \ref{fig:exp2_cvg_As}, or more explicitly in section \ref{sec:a1_d_cma_variants} of supplementary materials). We attribute this observation to the behavior sparsity of the task. CMA-ME is well known for pushing the search away from previously found solutions (\cite{fontaine2023covariance}). It limits CMA-ME from optimizing the performances of already-found solutions. CMA-MAE alleviates this issue with a rolling fitness threshold that conditions elites' insertion into the container. This constraint CMA-MAE's exploration capability, as it spends more budget in the same region of the search space. This mechanism allows CMA-MAE to better exploit previously found successes to find new ones on the most difficult task. On the allegro task, such a strategy might keep the evolutionary process to local minima. CMA-ME exploration capabilities help to escape those minima, while it can be detrimental in a more sparse behavioral domain.

The resulting QD-scores are provided in supplementary materials (section \ref{sec:a1_qd_scores}). Similarly to the first set of experiments (section \ref{sec:5_2_comparison_sota_qd_methods}), this metric is dominated by the coverage of $A_s$.

\begin{figure} %[t]
\begin{center}
\centerline{
 \includegraphics[width=1.0\textwidth]{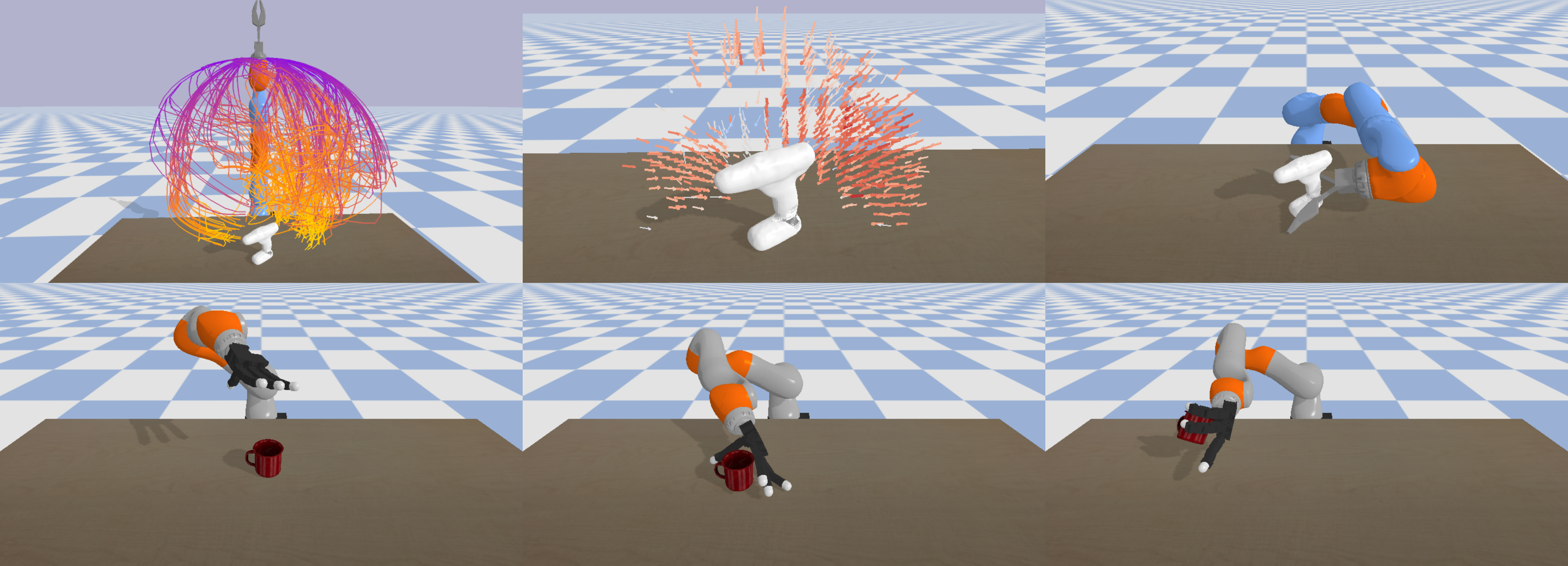}
}
\end{center}
\caption{\textbf{Output visualizations.} (Top left) 200 trajectories randomly sampled from an $A_s$ produced by ME-scs. Trajectories are displayed as a sequence of end effector positions. Color expresses temporality, from purple to yellow. The generated grasps are spread all over the operational space. (Top center) Visualization of the outcome descriptor associated with each solution from the same $A_s$. All the 724 solutions are displayed. The hotter the color, the higher the fitness. The regular space between the arrow's initial points expresses $A_s$ predefined sampling. The best-performing solutions are the ones that minimize the energy cost while maximizing grasp stability. (Top right) Grasping position corresponding to the best-performing individual from the same $A_s$ ($f=0.95$). (Bottom) A sequence of a randomly sampled success from a generated $A_s$ on the Allegro hand, from left to right.}
\label{fig:output_grasps_visualization}
\end{figure}

%%%%%%%%%%%%%%%%%%%%%%%%%%%%%%%%%%%%%%%%%%%%%%%%%%%%%%%%%%%%%%%%%%%%%%%
%                           6) Discussion
%%%%%%%%%%%%%%%%%%%%%%%%%%%%%%%%%%%%%%%%%%%%%%%%%%%%%%%%%%%%%%%%%%%%%%%

\section{Discussion}
\label{sec:6_discussion}

This section discusses the obtained results – with a spotlight on methods that were expected to get good performances on grasping. Section \ref{sec:6_1_top_perf_methods_vs_promising_methods} analyses the obtained results on RIBS for grasping. Section \ref{sec:6_2_detrimental_role_of_novelty_sparse_domains} provides a hypothesis on the detrimental role of novelty in sparse interaction domains, and section \ref{sec:6_3_eval_framework} discusses the proposed evaluation framework.

%=====================================================================%
%       6.1) Top performing methods vs promising methods
%=====================================================================%

\subsection{Top performing methods vs promising methods}
\label{sec:6_1_top_perf_methods_vs_promising_methods}

%--------------------------------------------------------------------%
%                     6.1.1) ME-scs
%--------------------------------------------------------------------%

\subsubsection{ME-scs}
\label{sec:6_1_1_me_scs}

The \textbf{high performances of ME-scs variants on grasping is the most striking result} obtained in the above-presented experiments. Grasping rewards greedy exploitation of previously found successes; focusing on local search from successful solutions significantly increases the probability of finding new ones. We can distinguish success-greedy (ME-scs-based) from fitness-greedy (ME-fit-based) variants, respectively focusing on finding new successes or optimizing the quality of already found ones. The gap in performances obtained with those methods compared to QD state-of-the-art shows that \textbf{there is room for developing QD algorithms that are suited to address tasks with similar properties}.

%--------------------------------------------------------------------%
%                     6.1.2) SERENE
%--------------------------------------------------------------------%

\subsubsection{SERENE}
\label{sec:6_1_2_serene}

One can easily create a domain that would show the limits of a ME-scs-like approach, by making the exploration-exploitation tradeoff proposed by SERENE a requirement for success. Such a domain would have the following characteristics: 1) the goal space $\mathcal{G}$ must consist of several distinct regions of $\mathcal{B}_{reach}$ that would require several mutations to move from one sub-region of $\mathcal{G}$ to another one, and 2) making each sub-region of $\mathcal{G}$ large enough to saturate ME-scs's population with solutions from a single region after having discovered it. Paolo et al.'s redundant arms or curling might be a good example of domains that satisfy those criteria. In this case, SERENE should outperform ME-scs, as its exploration-exploitation mechanisms would eventually lead to the discovery and exploration of each sub-regions of the goal space – while ME-scs would be trapped into the first found sub-region. It would confirm the difference in properties of those two algorithms: \textbf{SERENE balances exploration with exploitation}, which might block its execution into a standard NS algorithm if the rewarding regions are very hard to reach (as discussed in Paolo et al.), while \textbf{ME-scs explores until a reward zone is found}, and \textbf{then focuses on the exploitation of this first entry point} to generate as much diversity as possible while concurrently optimizing the reward of each rewarded solutions.

Beyond those theoretical questions, one might ask \textbf{what kind of non-toy problems can be addressed by each of those approaches}, considering the above-discussed properties. The present study considers the task of grasping, a key task for both concrete industrial cases (\cite{meszaros2022learning}) and open-ended scenarios in robotics (\cite{brohan2022rt}). The experimental results suggest that ME-scs's focus on exploitation is the most promising approach to generating a large repertoire of diverse high-performing grasping solutions. \textbf{Similar properties should arise in many other robotics manipulation tasks}, as most of them share some of the challenges studied in the present work. It includes behavioral sparsity, misalignment between the behavioral space and the targeted task, and hard-to-explore localized regions that concentrate the fitness function support.

%--------------------------------------------------------------------%
%                     6.1.3) CMA-MAE
%--------------------------------------------------------------------%

\subsubsection{CMA-MAE}
\label{sec:6_1_3_serene}

CMA-MAE (\cite{fontaine2023covariance}) is an extension of CMA-ME (\cite{fontaine2020covariance}) that alleviates some of its weaknesses – its inability to efficiently illuminate a behavior space when facing flat fitness landscapes. This CMA-ME limitation is visible here. While CMA-MAE outperforms most of the compared methods, \textbf{its performances are still way below ME-scs variants}. One might argue that \textbf{grasping rewards the greedy exploitation of previously found solutions}, as successful individuals lie in a limited part of the genotype space. CMA-MAE would therefore conduct a better exploration-exploitation compromise than a ME-scs method, resulting in lower grasping performances but better generalization capabilities on other tasks. This point can be nuanced by the fact that \textbf{grasping is itself submitted to discontinuities that require exploration to mutate one grasp to another successful trajectory that applies forces elsewhere on the object}. It might explain the difference in top-fitness performances for CMA-ME and CMA-MAE on the two tested robots. However, \textbf{we cannot state to what extent CMA-MAE struggles in tasks submitted to sparse behavior from those grasping experiments}. As the $\alpha$ parameter of CMA-MAE allows to control its tendency to explore or to behave similarly to a single-objective optimizer, it would be interesting to study how to make CMA-MAE more robust to sparse interaction problems. In particular, how to design QD algorithms that efficiently balance exploration and exploitation in those tasks, as looking for exploration might result in poor exploration.

%--------------------------------------------------------------------%
%                     6.1.4) NSMBS
%--------------------------------------------------------------------%

\subsubsection{NSMBS}
\label{sec:6_1_4_nsmbs}

NSMBS (\cite{morel2022automatic}) authors introduce their method as a way to address the challenges inherent to the task of grasping. The present work shows that NSMBS is not the only method that can efficiently address grasping. \textbf{Our results show that other methods – like ME-scs variants – perform way better on the task}. Indeed, NSMBS's ability to explore several behavior spaces during the evolutionary process has not been used. One might argue that: 1) making NSMBS evolve on multiple behavior spaces might lead to better results, and 2) that NSMBS's capacity to optimize several $\mathcal{B}_i$ allows producing an outcome repertoire with a high diversity on multiple components of the trajectory (e.g. how the end effector approaches the object, and how forces are applied on it). The visualization of individuals from a success archive produced by ME-scs (Figure \ref{fig:output_grasps_visualization}) shows that \textbf{choosing a single behavior space ($\bm{\phi_\mathcal{B}(\tau)=X_{a}^{touch}}$) does not result in limited diversity of grasping trajectory}. It is worth noting that the trajectories cover the whole operational space. Exploring the space of object-gripper contact points results in diverse ways to approach the object, as some opposite points on the object's surface are not likely to be reached with a similar approach – especially if we also optimize the energy consumption through fitness optimization. To avoid studying too many parameters simultaneously, we decided to let an in-depth analysis of multiBD on grasping for future work.

Identifying the best-performing method for doing RIBS on grasping is the main purpose of this paper. However, this study investigates a new kind of optimization problem – the sparse behavioral domains – in which some commonly admitted properties of QD algorithms do not hold. The most unexpected one is the detrimental role of novelty on the exploration capabilities of QD algorithms. This point is discussed in the following section.

\begin{figure} %[t]
\begin{center}
\centerline{
 \includegraphics[width=1.0\textwidth]{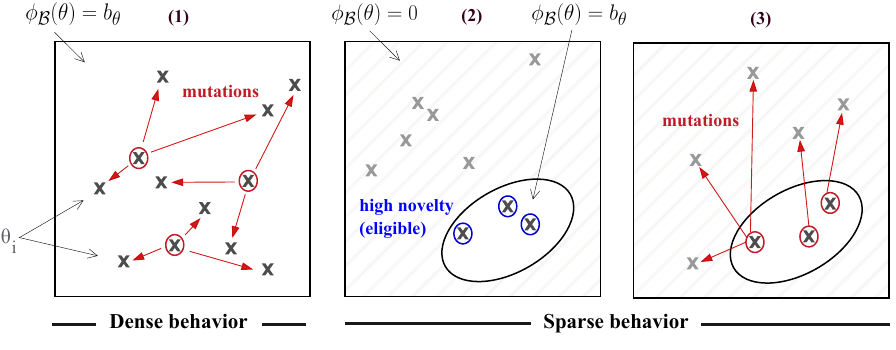}
}
\end{center}
\caption{\textbf{How novelty can be deceptive in sparse behavior domains.} (1) In dense behavior domains, novelty-guided optimization selects highly evolvable individuals that efficiently explore the behavior space. (2) When the behavioral landscape is flat, eligibility pushes the population toward the support of $\phi_\mathcal{B}$ through novelty-based selection. (3) Selection of highly evolvable individuals in sparse behavior domains can push the population out of the region of interest due to their high instability in $\mathcal{B}$. The novelty is here deceptive.
}
\label{fig:evolv_sparse_behavior}
\end{figure}

%=====================================================================%
% 6.2) On the detrimental role of novelty in sparse interaction domains
%=====================================================================%

\subsection{On the detrimental role of novelty in sparse interaction domains}
\label{sec:6_2_detrimental_role_of_novelty_sparse_domains}

Novelty-guided algorithms are well known for efficiently exploring a targeted behavior space (\cite{paolo2021sparse}). \textbf{Our results show that this property does not hold in sparse interaction context}, as NS explores similarly – or significantly worse – than quality-guided methods or a simple random search. We attribute this statement to the role of evolvability in novelty-guided methods. By applying pressure on novelty for selection, novelty-guided methods favor highly evolvable individuals (\cite{doncieux2020novelty}). \textbf{The selected solutions are the most behaviorally unstable ones, which are the more likely to fall into the non-eligible regions of the search space} (Figure \ref{fig:evolv_sparse_behavior}). This is why NS performs similarly to quality-guided methods on the problem submitted to strong behavioral sparsity (\textit{wsg50\_far}), while it reported the worst $cvg(A_o)$ performances on the less behavioral sparse problem (\textit{kuka\_close}).

The obtained results show that in sparse interaction context, \textbf{using methods that directly optimize $\bm{cvg(\mathcal{B})}$ is not the best strategy to optimize it}. As discussed above, \textbf{the novelty is here \textit{deceptive} regarding exploration}, as maximizing the evolvability is not the best strategy to explore a specific part of the targeted space in this context. Theoretical and experimental evidence regarding this specific phenomenon is left for future work. 

Note that this point is a good example of why QD practitioners should include QD methods of different nature in their experiments, as the above-stressed property is shared by NS and ME derivated methods that are guided by novelty. These comparisons have been made possible thanks to the proposed evaluation framework, which is discussed in the next section.

%=====================================================================%
% 6.3) On the evaluation framework
%=====================================================================%

\subsection{On the evaluation framework}
\label{sec:6_3_eval_framework}

This experiment shows that the proposed evaluation framework has many advantages for studying QD methods: it is \textbf{easy to implement, interpret and visualize}. By distinguishing the evaluation from the algorithmic internal components, \textbf{methods based on different kind of archives can fairly be compared}. Such an evaluation framework can similarly be used to compare methods with several behavioral spaces (\cite{kim2021exploration}, \cite{morel2022automatic}) with QD state-of-the-art approaches. It also has the benefit of \textbf{making the outcome of QD algorithms explicit} on a given problem. We hope this approach will be exploited in future QD works to allow more flexibility in algorithm design without compromising the experimental studies' interpretability or exhaustiveness.

A limitation of this evaluation is that it might favor ME-derivated variants – as the sampling within the container matches the sampling of the outcome archive. An output structured archive has however many benefits: most of the QD works focus on SA-based methods, and it allows easy visualization, analysis, and exploitation. Practitioners who want to have fine control over granularity according to expert knowledge can set a specific sampling (like in this work), or can rely on a CVT sampling to fix the maximal size of the outcome archive without having to design the grid cell sampling.

%%%%%%%%%%%%%%%%%%%%%%%%%%%%%%%%%%%%%%%%%%%%%%%%%%%%%%%%%%%%%%%%%%%%%%%
%                           7) Conclusion
%%%%%%%%%%%%%%%%%%%%%%%%%%%%%%%%%%%%%%%%%%%%%%%%%%%%%%%%%%%%%%%%%%%%%%%

\section{Conclusion}
\label{sec:7_conclusion}

This work investigates Quality-Diversity (QD) methods under sparse reward and sparse interactions problems through the application of grasping in robotics. What was arguably the most promising QD state-of-the-art methods for this domain (CMA-MAE, SERENE, and NSMBS) are significantly outperformed by variants of MAP-Elites that select non-null fitness solutions in priority. As a result, \textbf{an algorithm that is as simple as a standard MAP-Elites appeared to successfully generate a large set of diverse and high-performing solutions on multiple grasping domains}. 

The best-performing methods can thus be used to \textbf{automatically generate datasets of grasping trajectories}. As the access to demonstrations is a \textbf{key matter to solve grasping}, we believe that such dataset generators can provide significant help in the resolution of this task.

Our results suggest that addressing a task submitted to behavioral sparsity can lead to counterintuitive results. In particular, \textbf{explicitly guiding an algorithm to optimize the exploration of an outcome space favors selecting the most evolvable and unstable solutions, pushing the offspring out of the behavioral function support}. One might parallel seminal works in Novelty Search and deceptive reward, in which directly optimizing fitness might lead to poor optimization. In this context, \textbf{behavior sparsity results in deceptive novelty}. 

This paper opens many perspectives for the QD field regarding theoretical and practical matters: Do this work's observations hold in other robotic manipulation tasks? How to design more advanced algorithms that outperform ME-scs variants on grasping? How to avoid the sparse interaction setup through another behavioral characterization without compromising the algorithm's efficiency, as well as the output's interpretability and exploitability? More generally, we hope this work will incite the QD community on evolutionary robotics to tackle more complex problems, which are likely to result in breakthroughs in the field.

\section{Acknowledgement}

This work was supported by the Sorbonne Center for Artificial Intelligence, the German Ministry of Education and Research (BMBF) (01IS21080), and the French Agence Nationale de la Recherche (ANR) (ANR-21-FAI1-0004) - Learn2Grasp. It has received funding from the European Commission's Horizon Europe Framework Programme under grant agreement No 101070381 and from the European Union's Horizon Europe Framework Programme under grant agreement No 101070596. This work was performed using HPC resources from GENCI-IDRIS (Grant 20XX-AD011014320). Many thanks Emily Clement for her writing advices, and Alessia Loi, Charly Pecqueux-Guezenec, and Olivier Serris for their help and feedback.

\small

\bibliographystyle{apalike}
\bibliography{0_main}

\appendix

%%%%%%%%%%%%%%%%%%%%%%%%%%%%%%%%%%%%%%%%%%%%%%%%%%%%%%%%%%%%%%%%%%%%%%%
%                Supplementary materials
%%%%%%%%%%%%%%%%%%%%%%%%%%%%%%%%%%%%%%%%%%%%%%%%%%%%%%%%%%%%%%%%%%%%%%%

\clearpage
\begin{center}
\textbf{\large Supplementary Materials}
\end{center}

%=====================================================================%
%           A) Architecture of policies for grasping
%=====================================================================%

\section{Architecture of policies for grasping}
\label{sec:a1_architecture_of_policies_for_grasping}

QD domains for robotics have led to different architectures of policies: open-loop controllers (\cite{cully2015robots}), multiple layer perceptron (\cite{doncieux2019novelty}), or evolving neural networks (\cite{pugh2016quality}). Our controller for grasping consists of an open-loop trajectory guided by 3 waypoints. We initialize the gripper position to open and close it during the episode. Each genome follows the bellow described pattern:

\begin{equation*}
    \theta = \left(X_1, \alpha_1, X_2, \alpha_2, X_3, \alpha_3 \right)
\end{equation*}

Each $X_i=\left(x_i, y_i, z_i\right)$ coordinates define a waypoint in the cartesian space, and each $\alpha_i=(\alpha_i^p$, $\alpha_i^r$, $\alpha_i^y)$ values define the orientation the end effector must match at each waypoint in Euler angles with respect to the world basis. All those values are normalized to lie between $-1$ and $1$, according to the predefined limits of the operational space. To evaluate an individual, we first project back the normalized coordinates into the Cartesian space and then apply a polynomial interpolation such that each of the 3 points should be respectively reached at $T/3$, $2T/3$ and $T$ steps, where $T$ is the episode length. Each robot is initialized at a fixed position. 

When the end effector first touches the object, the gripper is closed with constant force. This mechanism is inspired by the \textit{Palmar Grasp Reflex}, which makes newborn infants closes their hands when pressure and touch are applied to the palm (\cite{futagi2012grasp}). It is also well known in the robotics litterature that non-zero-velocity grasps make the problem significantly more challenging (\cite{meszaros2022learning}).

While parallel grippers are controllable with a single value, dexterous hands provide more degrees of freedom. To exploit the dexterity of the Allegro hand without making the problem too complex, we have defined a set of grasp primitives that correspond to synergies that could be applied to a real Allegro hand. On the dexterous hand's domain, we have thus added a value $l_{gp}$ to each genome that describes the label of the grasp primitive. Details of the $n_{gp}$ designed primitives are provided in the hyperparameters section. The $\left [ -1,1 \right ]$ interval is uniformly sampled into $n_{gp}$ parts, such that $l_{gp}$ can be associated with a unique grasp primitive label. The corresponding grasp primitive is applied during the evaluation as soon as the object is first touched.

%=====================================================================%
%                        B) Fitness
%=====================================================================%

\section{Fitness}
\label{sec:a1_b_fitness}

To bypass the fitness sparsity of grasping, many works in robotics or RL relies on \textit{reward-shaping} to bootstrap the learning. It consists of manually designing a modular reward function to push the agent toward validating the sparse success criterion (\cite{lobbezoo2021reinforcement}). Many open-source environments for manipulation tasks in robotics provide a reward signal by default that makes the problem tractable with state-of-the-art methods. Consequently, the rare QD works on such domains deal with dense reward functions (\cite{salehi2022few}).

The main drawback of this approach is that it strongly biases how to solve the task. If we optimize a fitness function that rewards the agent when the object is grasped and when its end effector is getting closer to the object's center of mass, we do not learn "grasping" policies. Instead, we are addressing the derivated task: "grasp by applying forces around the object center of mass as fast as possible."

This is problematic for many reasons: firstly, it is unsatisfying from a learning point of view, as what we ultimately are interested in here is to make an agent learn to grasp; secondly, such a hand-designed fitness puts heavy constraints on the diversity of the generated solutions; finally, those works are not likely to result into interesting applications if the developed methods are not easily deployed on other sub-tasks one might want to address (e.g. grasping cautiously, grasping for a specific affordance), which is contradictory with the dependence to a certain way of solving the problem. 

Instead, we here condition the obtention of reward on the validation of the binary sparse success criteria. Nevertheless, we are interested in optimizing a fitness signal, as we here want to study grasping from a RIBS perspective. We have therefore designed the following fitness function:

\begin{equation*}
    f(\tau_i)= \left\{\begin{matrix}
0.5\times f_{ec}(\tau_i) + 0.5\times f_{gs}(\tau_i) \hspace{2mm} & \text{if \hspace{2mm} $f_c(\tau_i)=1$} \\
0 & \text{otherwise}
\end{matrix}\right.
\end{equation*}
where $f_c:S_\tau \rightarrow \left\{0,1 \right\}$ is the grasping binary \textit{success criterion}, $f_{ec}:S_\tau \rightarrow \mathbb{R}^+$ is the \textit{energy consumption fitness}, $f_{gs}:S_\tau \rightarrow \mathbb{R}^+$ is the \textit{grasp stability fitness}. $f_{ec}(\tau)$ and $f_{gs}(\tau)$ are normalized to lie in $\left [ 0, 0.5 \right ]$. In practice, $f_c$ returns 1 if the following conditions are verified for $N_g$ steps consecutively: 1) this object does not touch the table; 2) the object does not touch the floor; 3) the robot does not touch the table; 4) the end effector of the robot is touching the object; and 5) there is no penetration between the 3D models of the robot and the object. The energy consumption fitness is the sum of the applied torques on each joint throughout the episode. The grasp stability fitness is measured as follows:
\begin{equation*}
    f_{gs}(\tau_i) = (-1)* \left ( \frac{ var( \{ X_a^{touch_{i=1,...,T}} \} ) + var( \{ X_{obj}^{touch_{i=1,...,T}} \} ) }{n^{max\_touch} \times n^{cont\_touch} } + \zeta_d \right )
\end{equation*}
with $X_a^{touch_i}$ being the end effector-object contact point on the agent, $X_{obj}^{touch_i}$  being the end effector-object contact point on the object, $n^{max\_touch}$ being the number of iterations in which there has been contact between the agent and the object, $n^{cont\_touch}$ being the maximal number of iterations of continuous contact from the first touch between the agent and the object, and $\zeta_d$ being an additional cost that penalizes discontinuous interactions. The cost $\zeta_d$ is defined as the number of iterations from the first step at which the agent stops to touch the object until the end of the episode. The whole fitness is set as a negative function to maximize. The discontinuity cost is the dominating member of $f_{gs}$ until continuous grasps are found. As $f_{gs}$ cannot be lower than $1-T$ (meaning that the object is touched during the first step only) and larger than 0, we rely on the assumption that $f_{gs}\in \left [ -T, 0 \right ]$ to project it into the expected $\left [ 0, 0.5 \right ]$ interval. The normalization of $f_{ec}$ relies on extrema values measured from executions of fitness-free methods (Random and NS).

%=====================================================================%
%                        C) QD-scores
%=====================================================================%

\section{Qd-scores}
\label{sec:a1_qd_scores}

\begin{figure}[H] %[t]
\begin{center}
\centerline{
 \includegraphics[width=1.0\textwidth]{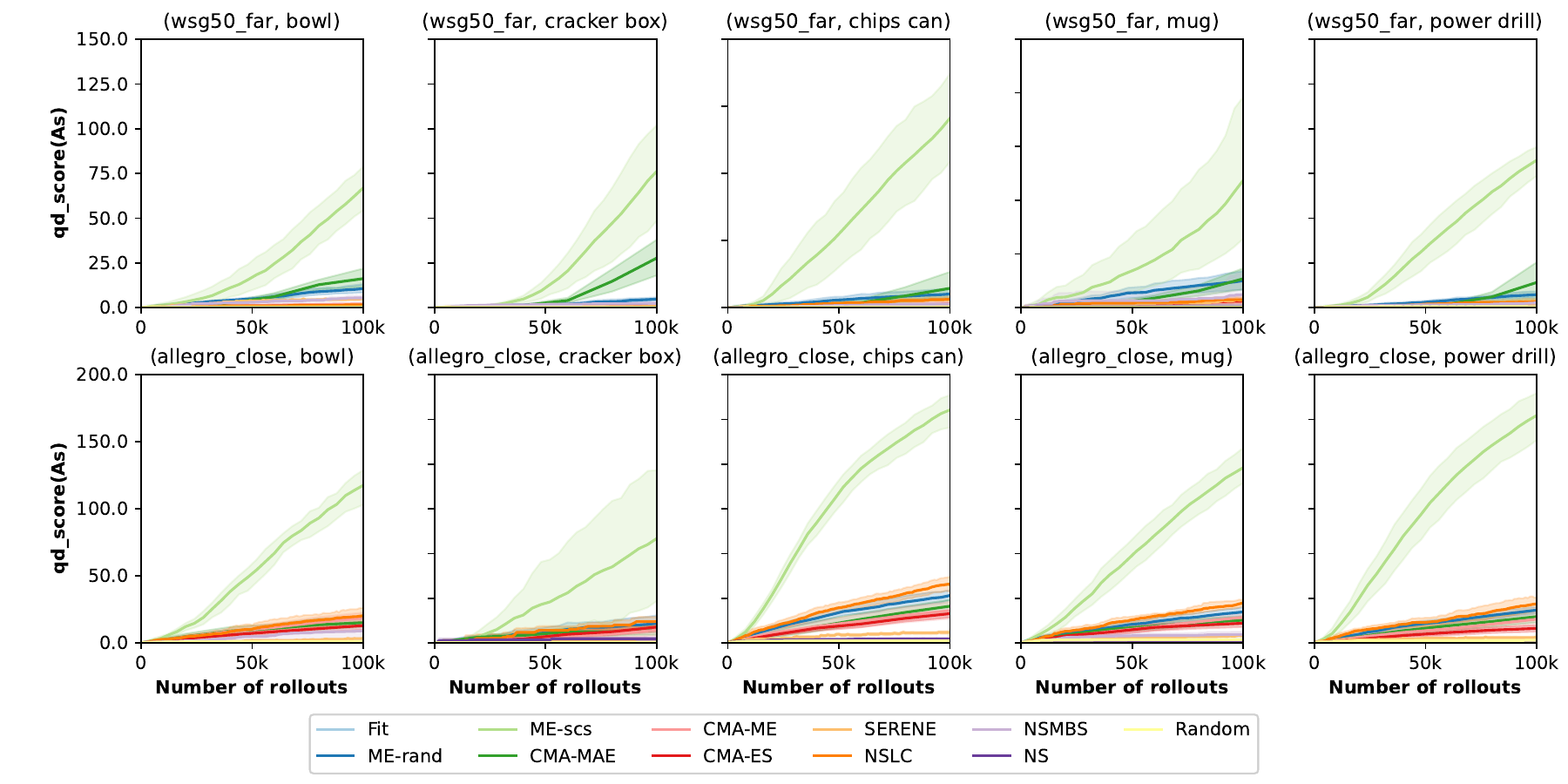}
}
\end{center}
\caption{\textbf{Qd-scores measured throughout the evolutionary process for state-of-the-art QD algorithms}. The large differences in success archive size make the qd-score saturated by the number of solutions. This measure does not provide information on the quality of the generated successes.}
\label{fig:a1_exp1_qd_score}
\end{figure}

\begin{figure}[H] %[t]
\begin{center}
\centerline{
 \includegraphics[width=1.0\textwidth]{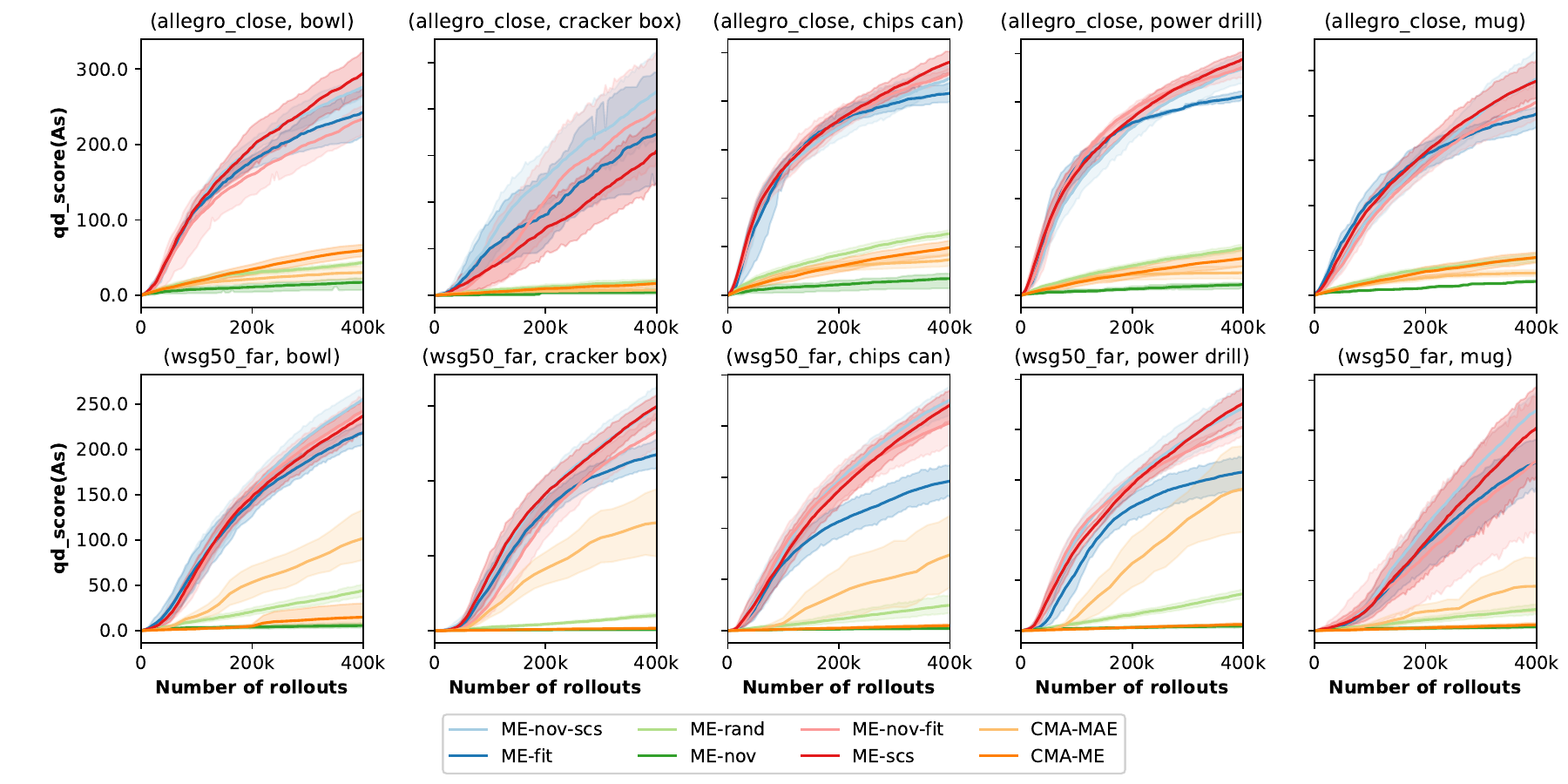}
}
\end{center}
\caption{\textbf{Qd-scores measured throughout the evolutionary process for different variants of MAP-Elites.} Results are essentially the same as those delivered by the coverage of $A_s$: fitness-guided ME-* variants can get stuck into local minima.}
\label{fig:a1_exp2_qd_score}
\end{figure}

%=====================================================================%
%           D) Success archive coverage of CMA-* variants
%=====================================================================%

\section{Success archive coverage of CMA-* variants}
\label{sec:a1_d_cma_variants}

\begin{figure}[H] %[t]
\begin{center}
\centerline{
 \includegraphics[width=1.0\textwidth]{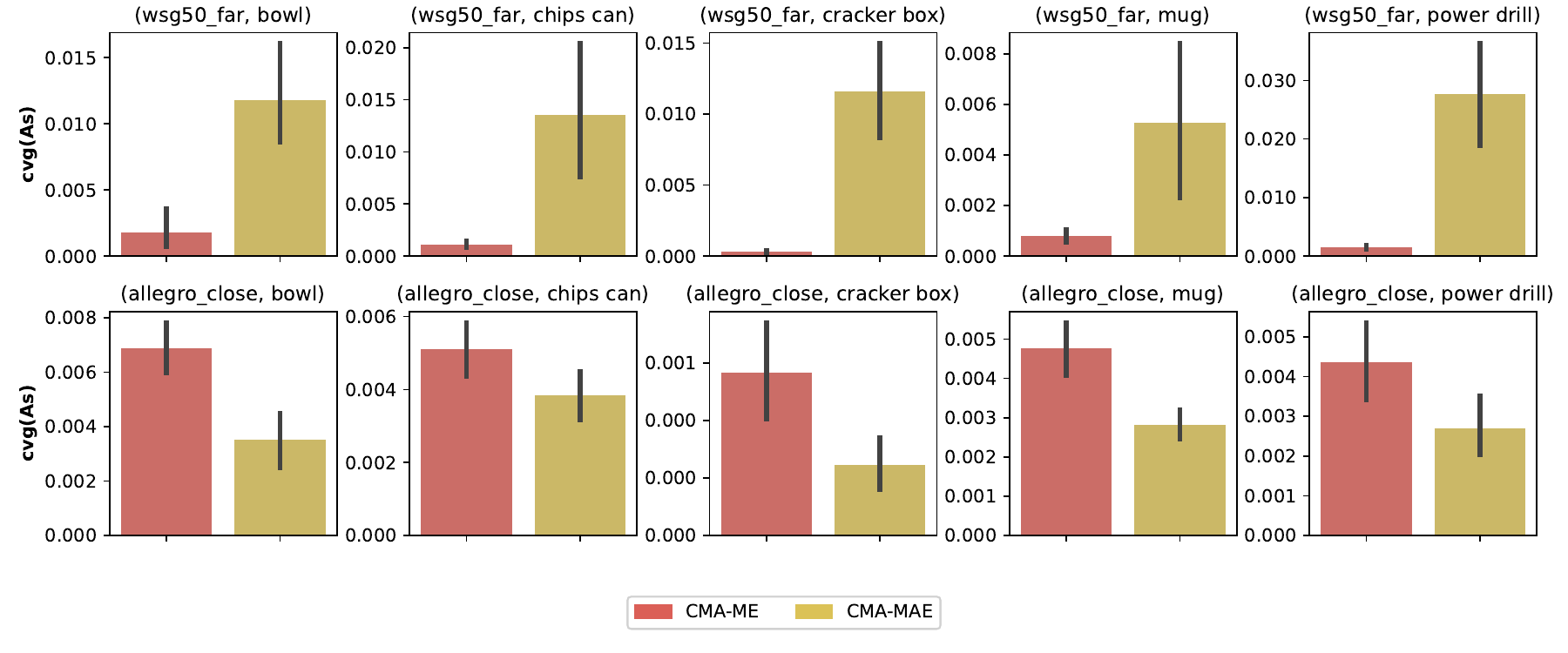}
}
\end{center}
\caption{\textbf{CMA-ME vs CMA-MAE on the coverage of the success archive} Over 10 seeds, after 400k evaluations. While CMA-MAE explores $\mathcal{G}$ more efficiently on the most difficult task (\textit{kuka\_wsg50\_far}), CMA-ME is better on the easier one (\textit{kuka\_allegro\_close}. We attribute this observation to the behavior sparsity of the task. CMA-ME is well known for pushing the search away from previously found solutions (\cite{fontaine2023covariance}). It limits CMA-ME from optimizing the performances of already-found solutions. CMA-MAE alleviates this issue with a rolling fitness threshold that conditions elites' insertion into the container. CMA-MAE spends more budget in the same region of the search space, limiting its exploration capabilities. This mechanism allows CMA-MAE to better exploit previously found successes to find new ones on the most difficult task. Such a strategy might keep the evolutionary process into local minima on the Allegro task. CMA-ME exploration capabilities help to escape those minima, while it can be detrimental under more sparse behavioral domains.}
\label{fig:a1_cvg_cma_me_supp_map}
\end{figure}

%\section{Introduction}
%\label{sec:1_introduction}

%\begin{figure}[H] %[t]
%\begin{center}
%\centerline{
% \includegraphics[width=1.0\textwidth]{assets/exp1/bds_study_outcome_archive_cvg_hist.pdf}
%}
%\end{center}
%\caption{\textbf{...} ...}
%\label{fig:a1_exp1_outcome_archive_hist_non_agg}
%\end{figure}

\end{document}